\documentclass[lettersize,journal]{IEEEtran}
\usepackage{amsmath,amsfonts}
\usepackage{array}
\usepackage[caption=false,font=normalsize,labelfont=sf,textfont=sf]{subfig}
\usepackage{textcomp}
\usepackage{stfloats}
\usepackage{url}
\usepackage{verbatim}
\usepackage{graphicx}
\usepackage{cite}

\usepackage{multirow} 
\usepackage{booktabs}
\usepackage{xcolor}
\usepackage{colortbl} 
\usepackage{hyperref}[colorlinks,linkcolor=blue] 
\usepackage[ruled,vlined,linesnumbered,lined,boxed,commentsnumbered]{algorithm2e}
\usepackage{algpseudocode}
\usepackage{subfloat} 

\begin{document}

\title{CoDS: Enhancing Collaborative Perception in \\Heterogeneous Scenarios via Domain Separation}

\author{Yushan Han, Hui Zhang,~\IEEEmembership{Member,~IEEE,} Honglei Zhang, \\Chuntao Ding,~\IEEEmembership{Member,~IEEE,} Yuanzhouhan Cao, Yidong Li,~\IEEEmembership{Senior Member,~IEEE} 
\thanks{This work was supported in part by the Fundamental Research Funds for the Central Universities 2023JBZY031 and 2025JBMC018, in part by the National Natural Science Foundation of China under Grant U2268203 and 62203040, and in part by the Beijing Natural Science Foundation L221011.}
\thanks{Yushan Han, Hui Zhang, Honglei Zhang, Yuanzhouhan Cao and Yidong Li are with Key Laboratory of Big Data \& Artificial Intelligence in Transportation (Beijing Jiaotong University), Ministry of Education and School of Computer Science and Technology, Beijing Jiaotong University, Beijing 100044, China.
E-mail: \{yushanhan, huizhang1, honglei.zhang, yzhcao, ydli\}@bjtu.edu.cn. (Corresponding author: \textit{Yidong Li}.)
}
\thanks{Chuntao Ding is with the School of Artificial Intelligence, Beijing Normal University, Beijing 100875, China (e-mail: ctding@bnu.edu.cn).}}

\markboth{Journal of \LaTeX\ Class Files,~Vol.~14, No.~8, August~2021}%
{Shell \MakeLowercase{\textit{et al.}}: A Sample Article Using IEEEtran.cls for IEEE Journals}


\maketitle

\begin{abstract}
  Collaborative perception has been proven to improve individual perception in autonomous driving through multi-agent interaction. Nevertheless, most methods often assume identical encoders for all agents, which does not hold true when these models are deployed in real-world applications. To realize collaborative perception in actual heterogeneous scenarios, existing methods usually align neighbor features to those of the ego vehicle, which is vulnerable to noise from domain gaps and thus fails to address feature discrepancies effectively. Moreover, they adopt transformer-based modules for domain adaptation, which causes the model inference inefficiency on mobile devices. To tackle these issues, we propose CoDS, a \underline{Co}llaborative perception method that leverages \underline{D}omain \underline{S}eparation to address feature discrepancies in heterogeneous scenarios. The CoDS employs two feature alignment modules, i.e., Lightweight Spatial-Channel Resizer (LSCR) and  Distribution Alignment via Domain Separation (DADS). Besides, it utilizes the Domain Alignment Mutual Information (DAMI) loss to ensure effective feature alignment. Specifically, the LSCR aligns the neighbor feature across spatial and channel dimensions using a lightweight convolutional layer. Subsequently, the DADS mitigates feature distribution discrepancy with encoder-specific and encoder-agnostic domain separation modules. The former removes domain-dependent information and the latter captures task-related information. During training, the DAMI loss maximizes the mutual information between aligned heterogeneous features to enhance the domain separation process. The CoDS employs a fully convolutional architecture, which ensures high inference efficiency. Extensive experiments 
  demonstrate that the CoDS effectively mitigates feature discrepancies in heterogeneous scenarios and achieves a trade-off between detection accuracy and inference efficiency.
\end{abstract}


\section{Introduction}

\begin{figure}[htbp]
  \centerline{\includegraphics[width=1\linewidth]{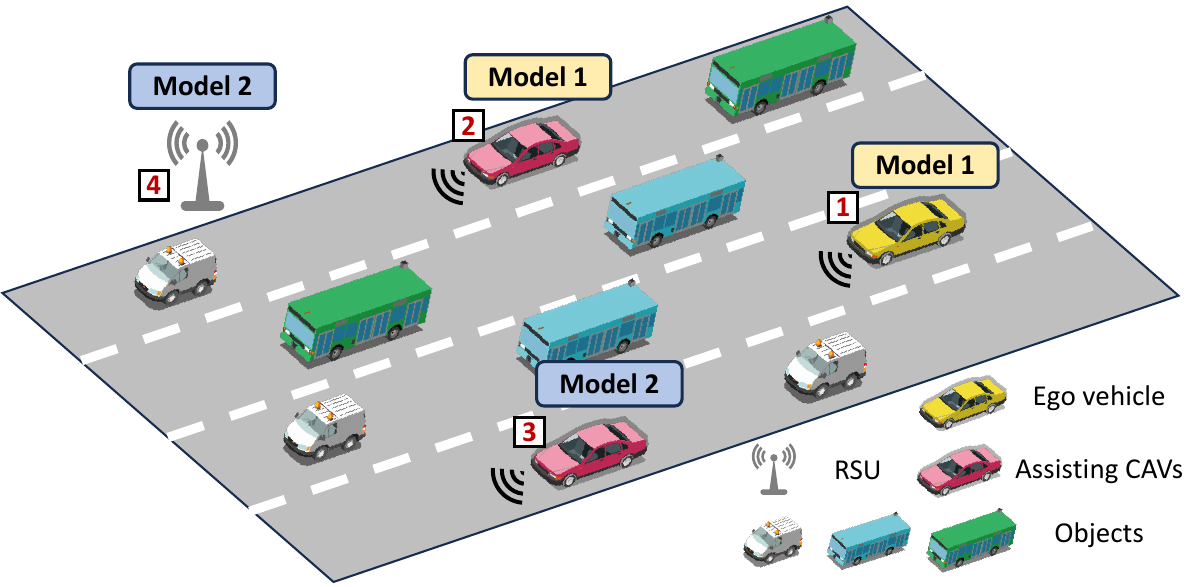}}
  \caption{\textbf{Comparison between homogeneous and heterogeneous scenarios}. In homogeneous scenario, Ego vehicle (No. 1) and CAV (No. 2) employ the identical model, resulting in shared homogeneous features. In heterogeneous scenario, Ego vehicle (No. 1) and CAV (No. 3) or RSU (No. 4) employ distinct models, leading to shared heterogeneous features.}
  \label{fig1}
  \vspace{-1em}
\end{figure}


\IEEEPARstart{A}{s} the number of vehicles continues to rise, the rapid advancement of intelligent transportation systems and autonomous driving technologies offers innovative solutions to address challenges in traffic efficiency and road safety. Among these, collaborative perception \cite{ye2023accuracy, pacp,xiao2023toward,zhou2024v2i,zhao2023meson,qian2024collaborative,wang2023ai, zhou2023dag,xie2024towards}, which leverages multi-agent interactions to overcome occlusion and long-range limitations faced by individual vehicles, has gained significant attention in the field of autonomous driving.
Collaborative perception usually includes vehicle-to-vehicle (V2V) \cite{xu2023v2v4real,OPV2V} and vehicle-to-everything (V2X) \cite{yu2022dair,V2X-VIT} modes, enabling interaction among ego vehicles, assisting connected autonomous vehicles (CAVs) and roadside units (RSUs).
Collaborative perception is categorized into early, intermediate and late collaboration, and most methods adopt intermediate collaboration for their fusion flexibility and low bandwidth requirements.
Recent studies concentrate on improving communication mechanisms \cite{where2comm,ERMVP,yang2023how2comm}, fusion strategies \cite{chen2023transiff,li2024di}, and mitigating the noise issue caused by communication latency \cite{wei2023asynchrony,yu2023flow} and localization errors \cite{coalign,feaco,roco}.  

Despite the notable success, existing methods often focus on homogeneous scenarios, where different agents employ identical encoders to extract features of the same size and distribution, thereby simplifying the feature fusion process. However, in practical applications, heterogeneous scenarios are more prevalent due to variations in hardware and software configurations \cite{feng2021review}. As shown in Fig.~\ref{fig1}, in heterogeneous scenarios, different encoders deployed in mobile devices extract features with discrepancies in both dimension and distribution, which can be attributed to distinct parameters and inductive biases \cite{hao2023one}. For example, 3D object detector encoders with varying architectures and hyperparameters exhibit different sensitivities to fine-grained details \cite{xu2023bridging}, resulting in feature misalignment. This misalignment poses challenges for directly applying existing feature fusion methods, leading to performance degradation in collaborative perception and potentially compromising traffic safety. Consequently, enabling connected autonomous vehicles to collaborate effectively in heterogeneous scenarios has become a signiﬁcant research area.

\begin{figure}[h]
  \centerline{\includegraphics[width=1\linewidth]{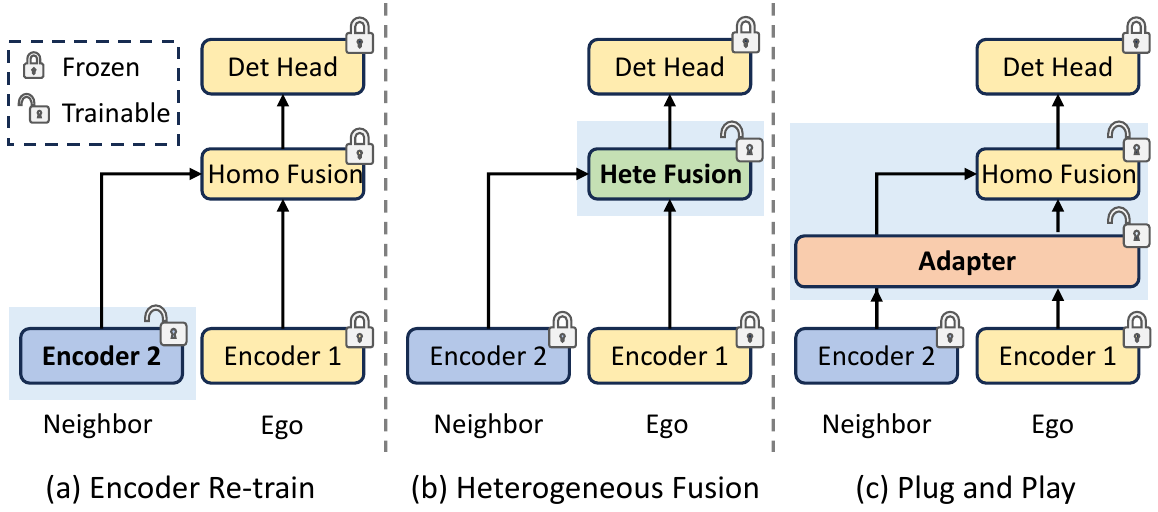}}
  \caption{\textbf{Different frameworks for heterogeneous collaboration}. (a) Re-trains encoders for neighbor agents, (b) designs a dedicated heterogeneous feature fusion module, and (c) incorporates a plug-and-play adapter that requires only fine-tuning based on the existing homogeneous feature fusion module.}
  \label{fig2}
  \vspace{-1em}
\end{figure}

Some methods have been proposed to address this issue, which can be classified into three categories (Fig.~\ref{fig2}).
(a) Encoder re-train: This type of work \cite{HEAL} retrains the encoders of neighbor agents to align them with the feature space of the ego vehicle. However, this approach requires access to the original encoder architectures of neighbor agents, which can be challenging when agents come from different companies.
(b) Heterogeneous fusion: These methods \cite{Hetecooper} design specialized modules to aggregate features with domain gaps. This approach requires training the new fusion module from scratch.
(c) Plug-and-play adapter: This category \cite{xu2023bridging, PnPDA, stamp, PolyInter} introduces adapters to align features from heterogeneous encoders. These adapters are non-intrusive to the original agent models and only require fine-tuning on trained homogeneous fusion modules, offering a more flexible solution.

This work focuses on adapter-based methods for their flexibility and scalability. Existing adapter-based approaches \cite{xu2023bridging, PnPDA, PolyInter} typically employ transformers to align the feature distribution from the neighbor to that of the ego vehicle. For instance, MPDA \cite{xu2023bridging} utilizes a cross-domain transformer to convert heterogeneous features from neighbor agents into the feature space of the ego vehicle. Similarly, PnPDA \cite{PnPDA} introduces a transformer-based semantic converter to transform the neighbor heterogeneous features into the ego semantic space. 
PolyInter \cite{PolyInter} employs a transformer-based interpreter to project neighbor features, guided by a general prompt and agent-specific prompts.
Despite their effectiveness, these methods suffer from two problems. (1) Since different encoders have varying characteristics and capabilities in feature extraction, the forced feature distribution conversion is vulnerable to noise from domain gaps \cite{DSN} and results in information loss. (2) Furthermore, the high computational cost of the transformer reduces model inference efficiency, which is crucial for autonomous driving applications. Consequently, these methods achieve suboptimal performance in actual deployment.

In heterogeneous scenarios, features obtained from different agents can be considered multiple views of the same scene, containing both task-related and encoder-specific (task-unrelated) information. According to the classic hypothesis \cite{CMC}, valuable information is the one that is shared across multiple views. Substantial evidence from cognitive science and neuroscience \cite{smith2005development,den2012prediction} also supports the idea that such view-invariant representations are encoded in the brain. Therefore, task-related information across heterogeneous features is valuable for collaborative perception, while encoder-specific information reflects the inductive bias of the encoder and hinders effective feature fusion.
To address distribution discrepancy issues in heterogeneous scenarios, we only need to capture task-related (domain-invariant) information between multiple agents while discarding encoder-specific (domain-specific) information. In other words, we aim to separate domain-invariant features from domain-specific ones.

Building on the above observations, we propose CoDS, a concise and effective collaborative perception method for heterogeneous scenarios. The core idea is to extract task-related information while eliminating encoder-specific information through domain separation. The CoDS comprises two alignment modules, i.e., the Lightweight Spatial-Channel Resizer (LSCR) and Distribution Alignment via Domain Separation (DADS). It also utilizes the Domain Alignment Mutual Information (DAMI) loss to enhance effective feature alignment.
Specifically, the LSCR aligns the spatial and channel dimensions of neighbor features using a lightweight convolutional layer. After that, the DADS performs domain separation through two mechanisms: an encoder-specific module that removes domain-dependent information and an encoder-agnostic module that extracts task-related (domain-invariant) information. To improve inference efficiency, we leverage convolutional layers for domain separation, utilizing their parameter-sharing and parallelization capabilities instead of transformers.
During training, the DAMI loss maximizes the mutual information between aligned heterogeneous features, ensuring that the features processed by LSCR and DADS preserve only task-related information while discarding encoder-specific content. This enhances both the robustness and effectiveness of feature alignment.
In summary, the main contributions are as follows:
\begin{itemize}
    \item We propose CoDS, a fully convolutional collaborative perception adapter designed to mitigate feature discrepancy issues in heterogeneous scenarios through domain separation. The CoDS enhances collaborative perception performance while ensuring high inference efficiency.
    \item The LSCR and DADS are proposed to align heterogeneous features. Specifically, the LSCR aligns neighbor features across spatial and channel dimensions using a lightweight convolutional layer. Subsequently, the DADS employs both encoder-specific and encoder-agnostic modules to remove domain-dependent information and capture task-related information effectively.
    \item The proposed DAMI loss enhances domain separation by maximizing the mutual information between aligned ego features and aligned neighbor features. This ensures that aligned features from multiple views preserve only task-related information in the current scene.
    \item Extensive experiments on three large-scale collaborative perception datasets (V2V4Real, OPV2V and V2XSet), three classic homogeneous feature fusion modules and five heterogeneous scenarios, demonstrate the superiority of the proposed CoDS in mitigating feature discrepancies while ensuring inference efficiency.
\end{itemize}

\section{Related Work}

\subsection{Collaborative Perception}
Collaborative perception \cite{han2023collaborative,zhang2022parallel,han2023ssc3od,han2025codts} is a vital technique in autonomous driving, enabling mobile vehicles to overcome the limitations of individual perception \cite{roh2024fast,li2023leovr,zhang2019mask} through multi-agent interaction.
It can be broadly categorized into three types based on the transmitted information: early fusion using raw point clouds, intermediate fusion using bird's-eye view (BEV) features, and late fusion using detection outputs.
Existing methods primarily focus on efficient communication \cite{where2comm,yang2023how2comm,ERMVP,MRCNet}, adaptive feature fusion \cite{hmvit,core,yang2023spatio,chen2023transiff}, and addressing challenges such as time delays \cite{SyncNet,wei2023asynchrony,yu2023flow} and localization errors \cite{coalign,feaco,roco}. However, most approaches assume that all agents use identical encoders, an assumption that is often unrealistic in practice.

\subsection{Collaborative Perception in Heterogeneous Scenarios}

In real-world applications, heterogeneous scenarios are more common, leading to feature discrepancies that hinder effective information fusion. 
To address this issue, HEAL \cite{HEAL} retrains the encoders of newly added agents to align with the ego domain. Hetecooper \cite{Hetecooper} introduces a heterogeneous feature fusion module that directly operates under heterogeneous settings.
Furthermore, some methods \cite{xu2023bridging,PnPDA,stamp,PolyInter} align heterogeneous features by introducing lightweight and flexible adapters without re-training the original encoders.
Specifically, MPDA \cite{xu2023bridging} uses a cross-domain transformer to project neighbor features into the ego feature domain. PnPDA \cite{PnPDA} introduces a semantic converter for feature alignment and a semantic enhancer to enhance ego features. 
STAMP \cite{stamp} first trains a protocol network to construct a unified semantic domain, then trains local adapters and reverters for feature alignment.
PolyInter \cite{PolyInter} uses an interpreter network to project neighbor features into the ego agent's semantic space, guided by a general prompt and agent-specific prompts for each newly added neighbor.
These adapter-based methods typically use transformers to align neighbor and ego feature distributions, but this forced conversion is prone to domain-gap noise and information loss, while the high computational cost of transformers hinders inference efficiency. To address this, we propose a fully convolutional adapter with domain separation to mitigate noise vulnerability.


\subsection{Domain Adaptation}
Domain adaptation \cite{oza2023unsupervised,chen2020domain,zhang2024scale} addresses domain shift challenges when transferring knowledge across different domains. It encompasses various approaches, including feature-based, instance-based and model-based adaptation. Among these, feature-based adaptation is the most widely used, focusing on identifying domain-invariant features through techniques such as discrepancy minimization, adversarial learning and feature reconstruction.
In heterogeneous collaborative perception, the use of distinct encoders introduces discrepancies in feature distributions, hindering effective feature fusion. To address this, MPDA \cite{xu2023bridging} leverages adversarial learning for domain adaptation. Specifically, it introduces a domain classifier tasked with distinguishing features from different domains, while the adapter aims to align features to confuse the domain classifier. This adversarial learning helps the adapter generate domain-invariant representations.
On the other hand, PnPDA \cite{PnPDA} employs contrastive learning to extract semantic information from heterogeneous features. It considers features of the same object in two feature maps as positive sample pairs and maximizes their semantic similarity.
Unlike previous work, we adopt mutual information maximization for domain adaptation. 

\subsection{Mutual Information Estimation}
Mutual Information (MI) is an information-theoretic measure that quantifies the dependency and shared information between two variables. Since true probability distributions are often unknown in real-world scenarios, recent studies have introduced neural networks for MI estimation. For instance, MINE \cite{MINE} and InfoNCE \cite{CPC} estimate MI by maximizing variational bounds, while Club \cite{club} takes an alternative approach by minimizing variational bounds.
Building on these works, some collaborative perception methods employ mutual information estimation for representation learning. For example, CRCNet \cite{CRCNet} minimizes mutual information between fused feature pairs to reduce information redundancy from different neighbor agents, while CMiMC \cite{su2024makes} maximizes mutual information between individual features and fused features to retain discriminative information from different views.
In contrast to these methods, our approach leverages mutual information maximization to achieve domain alignment between heterogeneous features, ensuring effective feature fusion in collaborative perception.

\begin{figure*}[htbp]
  \centerline{\includegraphics[width=0.9\linewidth]{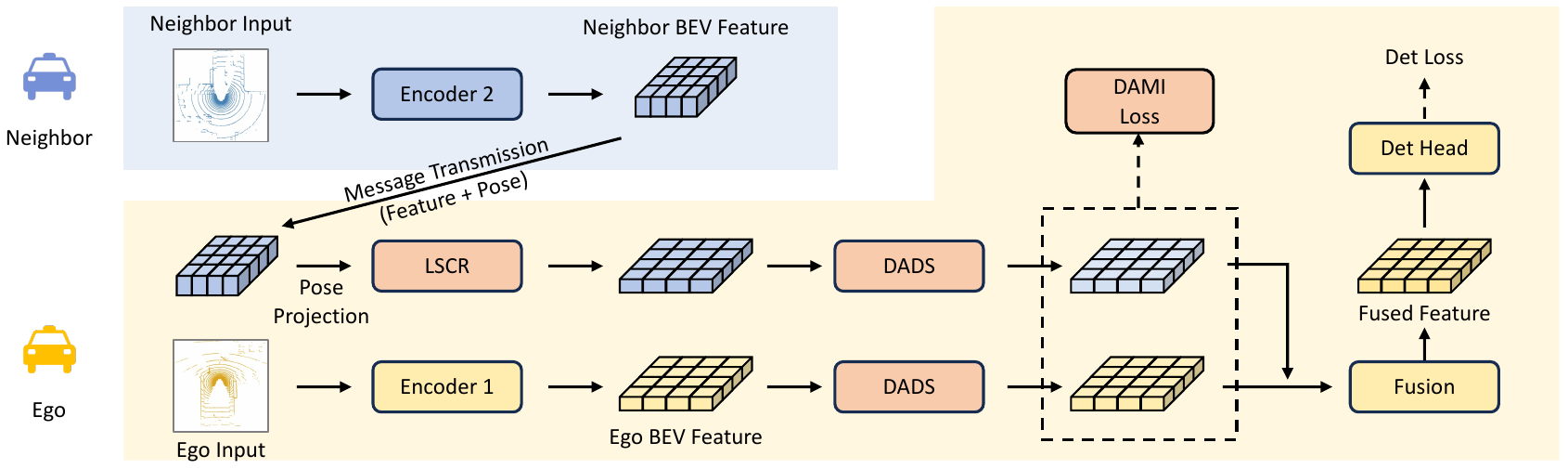}}
  \caption{\textbf{Overview of CoDS}. The ego and neighbor agents first extract features using distinct encoders. After receiving heterogeneous neighbor features, the ego applies the \textit{Lightweight Spatial-Channel Resizer} (LSCR) and \textit{Distribution Alignment via Domain Separation} (DADS) to align these features. During training, \textit{Domain Alignment Mutual Information} (DAMI) loss is used to ensure effective feature alignment.}
  \label{fig3}
  \vspace{-0.5em}
\end{figure*}

\section{Problem Formulation}
Consider $N$ agents in V2V or V2X collaboration perception scenarios. Let $x_i$ denote the LiDAR data collected by the $i$-th agent. The ego vehicle receives and fuses features from neighbor agents. The intermediate collaborative detection in a heterogeneous scenario can be formulated as:
\begin{align}
    f_{i}&=\boldsymbol{F}_{\text{enc}}^{\text{ego}}(x_{i}), \nonumber \\
    f_{j}&=\boldsymbol{F}_{\text{enc}}^{\text{nei}}(x_{j}), \nonumber \\
    f_{j\rightarrow i}&=\boldsymbol{F}_{\text{proj}}(\xi_{i},(f_{j},\xi_{j})),\nonumber\\
    \label{eq1} \widehat{f}_{i}&=\boldsymbol{F}_{\text{fuse}}(f_{i},f_{j\rightarrow i}), \nonumber\\
    y_{i}&=\boldsymbol{F}_{\text{\text{det}}}(\widehat{f}_{i}), 
\end{align}
where $\boldsymbol{F}_{\text{enc}}^{\text{ego}}$ and $\boldsymbol{F}_{\text{enc}}^{\text{nei}}$ denote encoders of ego vehicle and neighbor agents. $\boldsymbol{F}_{\text{proj}}$, $\boldsymbol{F}_{\text{fuse}} $ and $\boldsymbol{F}_{\text{\text{det}}}$ represent feature pose projection, feature fusion and detection head, respectively. And the ${\xi_i=(x_i,y_i,z_i, \theta_i, \phi_i, \psi_i)}$ is the 6-DoF pose of the $i$-th agent. The fused feature is denoted as $\widehat{f}_{i}$, and the collaborative detection output is denoted as $y_{i}$.

Training a collaborative detector is straightforward when encoders are identical, with features of the same size and distribution. However, in heterogeneous scenarios, where the encoders of the ego vehicle and neighbor agents differ, discrepancies in feature dimensions and distributions arise, leading to performance degradation after feature fusion. Our goal is to design a plug-and-play adapter to mitigate feature discrepancies while ensuring inference efficiency.

\section{Methodology}
This section proposes a collaborative perception method to alleviate the feature discrepancies in heterogeneous scenarios. We first introduce the framework, followed by the details of key modules and loss functions.

\subsection{Overall Architecture} \label{A}
To address the feature discrepancy issues in heterogeneous scenarios, we propose a collaborative perception method, called CoDS. As illustrated in Fig.~\ref{fig3}, the method comprises two alignment components and a loss function, i.e., LSCR module, DADS module and DAMI loss. i) The LSCR adjusts the size of neighbor features in both spatial and channel dimensions. ii) The DADS employs encoder-specific and encoder-agnostic domain separation modules to remove domain-dependent information and capture task-related information. iii) During training, the DAMI loss maximizes mutual information between aligned ego and neighbor features to ensure distribution alignment. Specifically, the proposed components are formulated as:
\begin{align}
    \bar{f}_{j\rightarrow i}&=\boldsymbol{F}_{\text{LSCR}}(f_{j\rightarrow i}),\nonumber\\
    \widetilde{f}_{i}, \widetilde{f}_{j\rightarrow i}&=\boldsymbol{F}_{\text{DADS}}(f_{i}, \bar{f}_{j\rightarrow i}).
\end{align}

Note that the distinct encoders are pre-trained in homogeneous scenarios and remain frozen in heterogeneous scenarios. Only the layers following the encoder are fine-tuned. To address the feature discrepancies issue, we employ the following steps before feature fusion for the ego vehicle. First, the LSCR module $\boldsymbol{F}_{\text{LSCR}}$ is employed to resize the projected neighbor features $f_{j\rightarrow i}$. Subsequently, the ego feature $f_{i}$ and the resized neighbor features $\bar{f}_{j\rightarrow i}$ are jointly passed through the DADS module $\boldsymbol{F}_{\text{DADS}}$, which effectively aligns their distributions. Finally, the aligned features $\widetilde{f}_{i}$ and $\widetilde{f}_{j\rightarrow i}$ are fused.

The proposed CoDS has the following advantages: i) We use domain separation modules to remove encoder-specific information and capture task-related information, thereby avoiding directly converting the domain of one encoder to another. ii) Benefiting from the parameter sharing and parallelization of the convolutional layer, our method is more efficient for training and inference than transformer-based collaborative perception methods.

\subsection{Lightweight Spatial-Channel Resizer} \label{B}


The Lightweight Spatial-Channel Resizer (LSCR) aims to adjust the neighbor features to align with the feature size of the ego vehicle. Given the ego feature $f_{i}\in\mathbb{R}^{H\times W\times C}$ and neighbor features $f_{j\rightarrow i}\in\mathbb{R}^{H^{\prime}\times W^{\prime}\times C^{\prime}}$, where $H^{\prime}\neq H, W^{\prime}\neq W, C^{\prime}\neq C$, the LSCR will adjust the neighbor features to $\bar{f}_{j\rightarrow i}=\boldsymbol{F}_{\text{LSCR}}(f_{j\rightarrow i})\in\mathbb{R}^{H\times W\times C}$ as follows:
\begin{align}
  f_{j\rightarrow i}^{0}&=\text{Conv}(f_{j\rightarrow i}), \nonumber\\
  \bar{f}_{j\rightarrow i}&=\text{BI}(f_{j\rightarrow i}^{0}), 
\end{align}
where we first apply $1\times1$ convolutional layers for channel alignment and get features $f_{j\rightarrow i}^{0}\in\mathbb{R}^{H^{\prime}\times W^{\prime}\times C}$. To achieve spatial alignment, we follow \cite{talebi2021learning} to adopt bilinear interpolation (BI) and obtain the resized features $\bar{f}_{j\rightarrow i}\in\mathbb{R}^{H\times W\times C}$.

\subsection{Distribution Alignment via Domain Separation} \label{C}
\begin{figure}[htbp]
  \centerline{\includegraphics[width=1\linewidth]{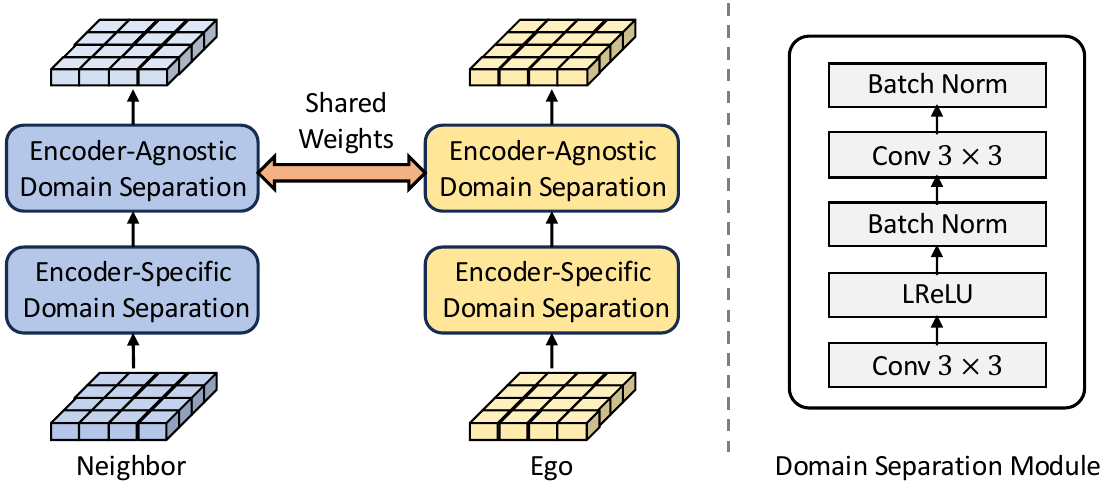}}
    \caption{\textbf{The architecture of the DADS}. 
    The encoder-specific domain separation modules remove private information tied to individual encoders, whereas the encoder-agnostic modules capture shared task-related information.}
    \label{fig5}
\end{figure}

\begin{figure*}[htbp]
  \centerline{\includegraphics[width=0.9\linewidth]{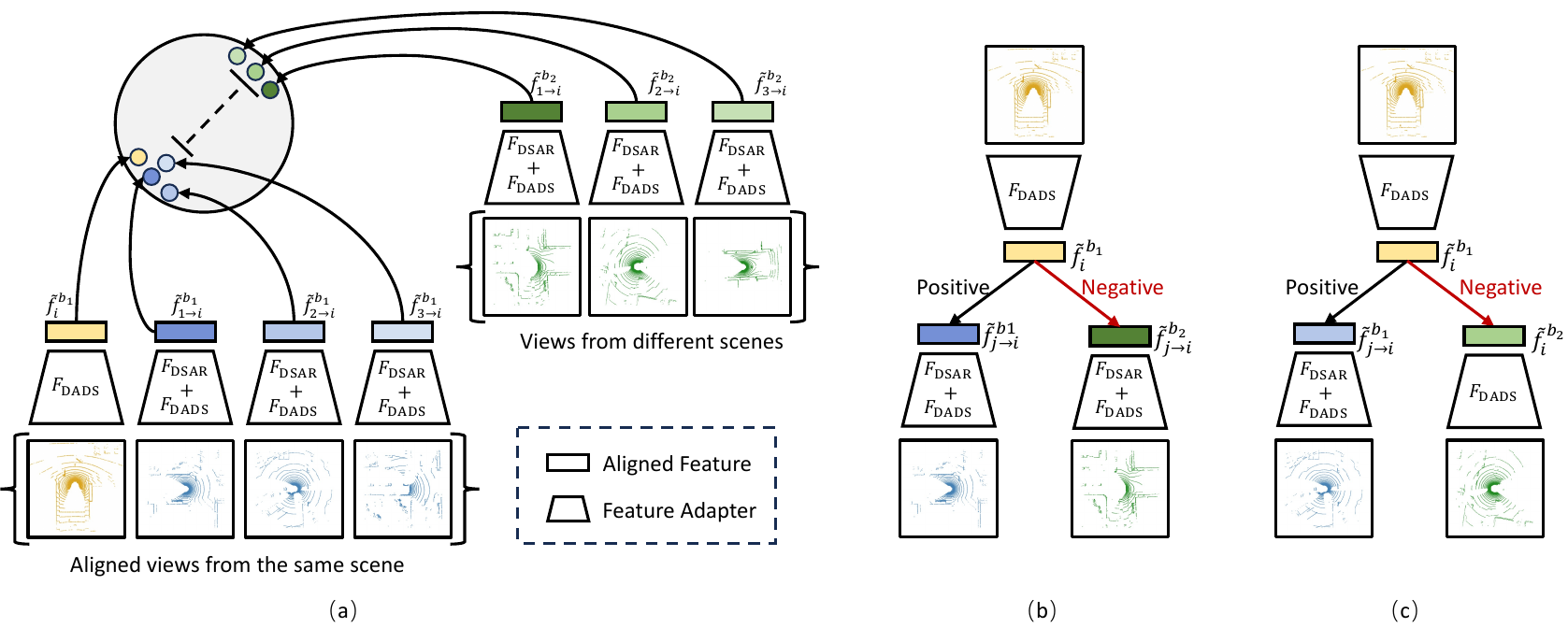}}
  \caption{\textbf{Illustration of the DAMI loss}. (a) We maximize mutual information between aligned features by bringing views of the same scene closer together and pushing views of different scenes further apart, which is achieved through contrastive learning on positive and negative pairs.
  (b) and (c) illustrate the construction of positive and negative samples. In scene $b_1$, the $j$-th aligned neighbor feature $\widetilde{f}^{b_1}_{j\rightarrow i}$ serves as a positive sample for the aligned ego feature $\widetilde{f}^{b_1}_{i}$. (b) If the ego in another scene $b_2$ also has the $j$-th agent, the negative sample is the aligned neighbor feature $\widetilde{f}^{b_2}_{j\rightarrow i}$ from scene $b_2$. (c) Otherwise, the negative sample is the aligned ego feature $\widetilde{f}^{b_2}_{i}$ from scene $b_2$.}
  \label{fig7}
  \vspace{-0.5em}
\end{figure*}

Given features $f_{i},\bar{f}_{j\rightarrow i}\in\mathbb{R}^{H\times W\times C}$ with distribution discrepancy, we denote their marginal distributions as $\mathbb{P}(f_{i})$ and $\mathbb{P}(\bar{f}_{j\rightarrow i})$, where $\mathbb{P}(f_{i})\neq\mathbb{P}(\bar{f}_{j\rightarrow i})$. Extensive research \cite{yao2021multisource,zhang2021c2fda} has demonstrated the existence of projection functions in domain adaptation, which effectively maps features from disparate distributions into a common space. Therefore, we propose the Distribution Alignment via Domain Separation (DADS), which employs projection functions for each domain, denoted as $\boldsymbol{M}_{\text{ego}}(\cdot)$ and $\boldsymbol{M}_{\text{nei}}(\cdot)$. These functions ensure that the projected features maintain similar marginal distributions, i.e.,$\mathbb{P}(\boldsymbol{M}_{\text{ego}}(f_i))\approx\mathbb{P}(\boldsymbol{M}_{\text{nei}}(\bar{f}_{j\rightarrow i}))$. 
 

As shown in Fig.~\ref{fig5}, the projection functions $\boldsymbol{M}_{\text{ego}}(\cdot)$ and $\boldsymbol{M}_{\text{nei}}(\cdot)$ are implemented using two types of domain separation modules. The first type is the encoder-specific domain separation modules $\boldsymbol{M}^{\text{es}}(\cdot)$, which are constructed independently for each domain and directly achieve domain separation by removing domain-dependent information. The second type is the encoder-agnostic domain-separation modules $\boldsymbol{M}^{\text{ea}}(\cdot)$, which employ a weight-sharing scheme and indirectly achieve domain separation by capturing task-related information (i.e., domain-invariant features) through projection into a common latent feature space. Both types of modules are arranged sequentially within the domain and share an identical structure, comprising two $3\times3$ convolutional layers, Batch Normalization, and the LeakyReLU activation function. Consequently, the overall projection functions can be expressed as:
\begin{align}
    \boldsymbol{M}_{\text{ego}}(\cdot)=(\boldsymbol{M}^{\text{es}}_{\text{ego}}\circ \boldsymbol{M}^{\text{ea}}_{\text{ego}})(\cdot), \nonumber \\
    \boldsymbol{M}_{\text{nei}}(\cdot)=(\boldsymbol{M}^{\text{es}}_{\text{nei}}\circ \boldsymbol{M}^{\text{ea}}_{\text{nei}})(\cdot), 
\end{align}
where $\circ$ denotes the connection of convolution layers. And features processed by domain separation modules can be expressed as $\widetilde{f}_{i}=\boldsymbol{M}_{\text{ego}}(f_{i})$ and $\widetilde{f}_{j\rightarrow i}=\boldsymbol{M}_{\text{nei}}(\bar{f}_{j\rightarrow i})$.


Note that both encoder-specific and encoder-agnostic domain separation modules are indispensable. Using only encoder-specific modules would preserve some private information tied to individual encoders, which hinders complete distribution alignment. Conversely, relying solely on encoder-agnostic modules would be vulnerable to encoder-specific information, thereby impeding the projection of features into a shared space. Further analysis and discussion can be found in the ablation study section.

\subsection{Domain Alignment Mutual Information Loss} \label{D}

In collaborative perception, heterogeneous features from different agents can be viewed as multiple views of the same scene, with only task-related information accurately representing the environment.
To ensure the adapter captures this task-related information while eliminating encoder-specific details, we maximize the mutual information (MI) between aligned ego and neighbor features. This enhances representation consistency across different views and effectively mitigates distribution discrepancies.
The MI between aligned ego feature $\widetilde{f}_{i}$ and aligned neighbor feature $\widetilde{f}_{j\rightarrow i}$ is defined as:
\begin{align}
  \mathcal{I}(\widetilde{f}_{i} ; \widetilde{f}_{j\rightarrow i})=\sum_{x \in \widetilde{f}_{i}} \sum_{y \in \widetilde{f}_{j\rightarrow i}} p(x, y) \log \frac{p(x, y)}{p(x) p(y)}.
\end{align}

Vanilla MI only measures the dependency between two random variables, which is insufficient for capturing dependencies across aligned features from multiple views. To fill this gap, we define the Domain Alignment Mutual Information (DAMI), which is adaptable to multiple views. Specifically, DAMI first constructs pairwise MI between the aligned ego feature $\widetilde{f}_{i}$ and each aligned neighbor feature $\widetilde{f}_{j\rightarrow i}$, then averages these pairwise MIs to form DAMI. Let $N_{\text{nei}}$ represent the total number of neighbor agents for the $i$-th ego vehicle. Then, the DAMI for the $i$-th ego can be formulated as:
\begin{align}
  \mathcal{I}_{\text{DAMI}}=\frac{1}{N_{\text{nei}}} \sum_{j=1}^{N_{\text{nei}}} \mathcal{I}(\widetilde{f}_{i} ; \widetilde{f}_{j\rightarrow i}).
\end{align}

\begin{figure}[h]
  \centerline{\includegraphics[width=1\linewidth]{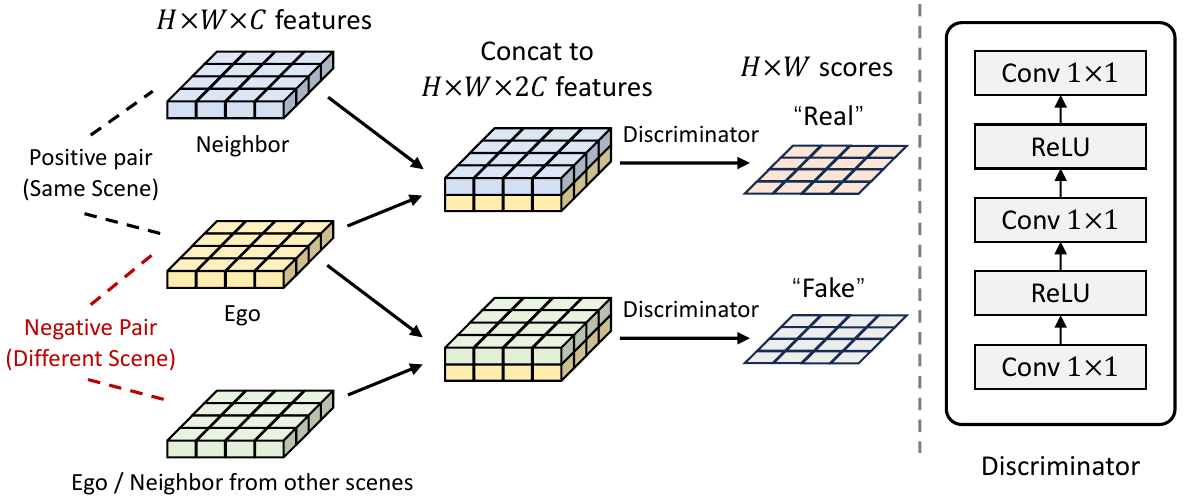}}
  \caption{\textbf{The architecture of discriminator}. Positive and negative feature pairs are concatenated separately, and a convolutional discriminator is then applied to score these “real” (positive) and “fake” (negative) feature pairs.}
  \label{fig8}
  \vspace{-0.5em}
\end{figure}

To mitigate distribution discrepancies between heterogeneous features, the CoDS aims to maximize DAMI, which requires a lower-bound estimation. Following \cite{CPC}, we estimate this MI lower bound using contrastive loss between aligned feature pairs, which can be formulated as:
\begin{align}
  \mathcal{I}(\widetilde{f}_{i} ; \widetilde{f}_{j\rightarrow i})\geq \log (k)-\mathcal{L}_{\text {contrast}}=\mathcal{\hat{I}}(\widetilde{f}_{i} ; \widetilde{f}_{j\rightarrow i}),
\end{align}
where the contrastive loss $\mathcal{L}_{\text{contrast}}$ is used to train a discriminator that distinguishes aligned features from different scenes. 
Specifically, the discriminator minimizes the loss by assigning a high score to positive pairs (i.e., aligned ego and neighbor features from the same scene) and a low score to negative pairs (i.e., aligned ego and neighbor features from different scenes). The parameter $k$ represents the number of negative pairs in the sample set, indicating that incorporating more negative samples can enhance the learning of the feature adapter.

Minimizing the objective $\mathcal{L}_{\text {contrast}}$ effectively maximizes the lower bound on the mutual information $\mathcal{\hat{I}}(\widetilde{f}_{i} ; \widetilde{f}_{j\rightarrow i})$. Therefore, the task of DAMI maximization is transformed into a contrastive learning minimization problem, and we can use $\mathcal{L}_{\text {contrast}}$ to represent the DAMI loss $\mathcal{L}_{\text{DAMI}}$:
\begin{align}
  \mathcal{L}_{\text{DAMI}} &=\max \frac{1}{N_{\text{nei}}} \sum_{j=1}^{N_{\text{nei}}}\mathcal{\hat{I}}(\widetilde{f}_{i};\widetilde{f}_{j\rightarrow i}) \nonumber\\
  &=\max \frac{1}{N_{\text{nei}}} \sum_{j=1}^{N_{\text{nei}}}\left[\log (k)-\mathcal{L}_{\text {contrast}}\right] \nonumber\\
  &=\min \frac{1}{N_{\text{nei}}} \sum_{j=1}^{N_{\text{nei}}}\left[\mathcal{L}_{\text {contrast}-\log (k)}\right]. 
\end{align}

\begin{algorithm}
  \SetKwData{Left}{left}\SetKwData{This}{this}\SetKwData{Up}{up}
  \SetKwFunction{Union}{Union}\SetKwFunction{FindCompress}{FindCompress}
  \SetKwInOut{Input}{Input}\SetKwInOut{Output}{Output}
  \Input{Aligned ego feature $\widetilde{f}_{i}$ and neighbor features $\{\widetilde{f}_{j\rightarrow i}\}_{j= 1}^{N_{\text{nei}}}$ in one iteration. Discriminator $D$.}
  \BlankLine
  $\text{ego\_anchor} \gets [\ ]$ \; 
  $\text{pos\_sample} \gets [\ ]$ \; 
  $\text{pos\_num} \gets [\ ]$ \;
  \tcp{\textcolor{blue}{Traverse all scenes in current iteration}}
  \For{{$b = b_1, b_2, \dots, b_n$}}{ 
    $\text{pos\_num} \gets N^b_{\text{nei}}$ \;
    $\text{ego\_anchor} \gets \widetilde{f}_{i}^b$ \;
    $\text{pos\_sample} \gets \{\widetilde{f}^b_{j\rightarrow i}\}_{j= 1}^{N^b_{\text{nei}}}$ \;
  }
  $\mathcal{L}_{cont} \gets [\ ] $\;
  \For{{$k = 1, 2, \dots, max(\textup{pos\_num})$}}{
    $\text{ego\_anchor\_k} \gets [\ ]$ \;
    $\text{pos\_sample\_k} \gets [\ ]$ \;
    $\text{neg\_sample\_k} \gets [\ ]$ \;
    \tcp{\textcolor{blue}{The $k$-th positive sample pairs}}
    \For{{$v = 1, 2, \dots, len(\textup{pos\_num})$}}{
      \If{$\textup{pos\_num}[v] \textgreater k$}{
        $\text{ego\_anchor\_k} \gets \text{ego\_anchor}[v]$ \;
        $\text{pos\_sample\_k} \gets \text{pos\_sample}[v][k]$\;
      }
    }
    \tcp{\textcolor{blue}{Find the $k$-th negative sample}}
    \If{$len(\textup{pos\_sample\_k}) \textgreater 1$}{
      \For{{$p = 2, \dots, len(\textup{pos\_sample\_k})$}}{
        $\text{neg\_sample\_k} \gets \text{pos\_sample\_k}[p]$
      }
      $\text{neg\_sample\_k} \gets \text{pos\_sample\_k}[1]$ \;
    }
    \Else{
      \Repeat{\( q \neq v \)}{
        \( q \leftarrow \text{Random}([1, len($\text{ego\_anchor}$)])\) \;
      }
      $\text{neg\_sample\_k} \gets \text{ego\_anchor}[q]$ \;
    }
    \tcp{\textcolor{blue}{Calculate contrastive loss}}
    $\mathcal{L}_{cont}\gets D(\text{ego\_anchor\_k}, \text{pos\_sample\_k}, \text{neg\_sample\_k})$\;
  }
  $\mathcal{L}_{\text {contrast}} \gets mean(\mathcal{L}_{cont})$\;
    \Output{$\mathcal{L}_{\text {contrast}}$ in current iteration}
  \caption{Calculate contrastive loss $\mathcal{L}_{\text {contrast}}$} \label{cont}
\end{algorithm}

To implement contrastive loss, we first construct positive and negative pairs. As shown in Fig.~\ref{fig7}(a), to achieve effective domain separation and enhance representation consistency across heterogeneous features, we treat aligned ego and neighbor features from the same scene as positive pairs, while aligned ego features and those from other scenes are treated as negative pairs. In each training iteration, we sample $B$ scenes, denoted as $\mathcal{B} = \{ b_1, b_2, \dots, b_n \} $. In a scene $b_1$, the aligned ego feature serves as the anchor, with each aligned neighbor feature as a positive sample for the anchor. As shown in Fig.~\ref{fig7}(b), for the $j$-th aligned neighbor feature, the negative sample is the $j$-th aligned neighbor feature from another scene  $b_2$. The different scenes and distinct encoders between the anchor and the negative sample ensure that negative pairs are pushed apart. However, not every collaborative scene has $j$ neighbor agents. To address this issue, we use the aligned ego feature as the negative sample when other scenes lack sufficient neighbor agents, as illustrated in Fig.~\ref{fig7}(c).
The structure of the discriminator is shown in Fig.~\ref{fig8}. For each positive and negative feature pair, we first concatenate the anchor with the positive or negative sample along the channel dimension. The concatenated features are then fed into the discriminator, which outputs  $H \times W$  score maps used to calculate the $\mathcal{L}_{\text {contrast}}$. The details of constructing positive and negative sample pairs and calculating contrastive loss are illustrated in Algorithm \ref{cont}.

\subsection{Overall Loss Function} \label{E}
The total loss $\mathcal{L}$ for training CoDS is summarized as follows:
\begin{align}
    \mathcal{L}&=\mathcal{L}_{\text{det}}+\beta_{\text{DAMI}}\mathcal{L}_{\text{DAMI}},\nonumber\\
    \mathcal{L}_{\text{det}}&=\alpha_{\text{cls}}\mathcal{L}_{\text{cls}}+\alpha_{\text{reg}}\mathcal{L}_{\text{reg}}+\alpha_{\text{dir}}\mathcal{L}_{\text{dir}},\nonumber
\end{align}
where $\mathcal{L}_{\text{det}}$ denotes detection loss, which includes focal loss for classification $\mathcal{L}_{\text{cls}}$, smooth-L1 loss for regression $\mathcal{L}_{\text{reg}}$ and softmax classification loss for direction $\mathcal{L}_{\text{dir}}$. Specifically, $\beta_{\text{DAMI}}$, $\alpha_{\text{cls}}$, $\alpha_{\text{reg}}$ and $\alpha_{\text{dir}}$ are the weights for loss functions. 


\section{Experimental Results}

\subsection{Datasets and Evaluation Metrics}

We validate the proposed CoDS method on the task of LiDAR-based collaborative 3D object detection using three large-scale collaborative perception datasets.


\textbf{V2V4Real} \cite{xu2023v2v4real} is the first large-scale, real-world V2V dataset collected using two vehicles. It includes 20K frames of point clouds, with 6,958 frames for training, 1,993 for validation, and 1,993 for testing.

\textbf{OPV2V} \cite{OPV2V} is a simulated V2V perception dataset where 2 to 7 collaborative vehicles, each equipped with a LiDAR and four cameras. It comprises 11,464 frames of 3D point clouds and 230K annotated 3D boxes, split into 6,374 training frames, 1,980 validation frames and 2,170 testing frames.

\textbf{V2XSet} \cite{V2X-VIT} is another simulated dataset designed for V2X applications, featuring both roadside units and autonomous vehicles. It contains 6,694 training frames, 1,920 validation frames and 2,834 testing frames.

We evaluate different methods on the testing sets of three datasets, assessing both accuracy and efficiency. For accuracy, we use 3D detection performance measured by average precision (AP) at Intersection-over-Union (IoU) thresholds of 0.50 and 0.70. For efficiency, we evaluate the frames per second (FPS) to measure the processing speed of models.

\subsection{Experimental Setups}
\subsubsection{Implementation Details}
The collaborative detector in the heterogeneous scenario is fine-tuned from the homogeneous scenario.
Initially, we train collaborative detectors with various encoders in the homogeneous scenario. Next, we load and freeze distinct pre-trained encoders for different agents and fine-tune adapters and fusion modules. 
This work simplifies the setting and only considers two different encoders.
To train the model, we use the Adam optimizer with a learning rate of 0.002. The weight of the DAMI loss is set to $\beta_{\text{DAMI}}=1$. For the detection loss, we adopt the same weight settings as PointPillars \cite{pointpillars}: $\alpha_{\text{cls}}=1$, $\alpha_{\text{reg}}=2$ and $\alpha_{\text{dir}}=0.2$.
All models are trained on NVIDIA RTX 4090.
For quantitative comparison, we select three classic feature fusion modules, Attfusion \cite{OPV2V}, DiscoNet \cite{DiscoNet} (student model only) and CoBEVT \cite{xu2022cobevt}. Specifically, DiscoNet \cite{DiscoNet} is selected to assess efficiency performance and to conduct ablation studies.



\subsubsection{Distinct Encoder} We select PointPillars \cite{pointpillars}, SECOND \cite{second} and VoxelNet \cite{voxelnet} as the detection encoders. 
The half LiDAR range (X$\&$Y), voxel resolution of the encoders, and feature size (C$\times$H$\times$W) are summarized in Table~\ref{tab:setting}.
The detection accuracy of collaborative detectors in homogeneous scenarios is listed in Table~\ref{tab:homo}.
In the subsequent experiments, $p_0$ and $p_1$ represent PointPillars with different voxel parameters. Similarly, $s_1$ and $v_1$ refer to SECOND and VoxelNet, respectively. We consider two heterogeneous settings:  i) The ego agent is equipped with the pre-trained $p_0$, while neighbor agents are equipped with pre-trained $p_1$, $s_1$ or $v_1$. ii) The ego agent is equipped with the pre-trained $s_1$ or $v_1$, while neighbor agents are equipped with pre-trained $p_0$.

\begin{table}[htbp]
  \Huge
  \renewcommand\arraystretch{1.3} 
  \centering 
  \caption{Detailed parameters of heterogeneous encoders.}
  \resizebox{1\linewidth}{!}{
  \begin{tabular}{c|c|c|c|c|c|c}
    \toprule
    \multicolumn{4}{c}{Encoder Setting} & \multicolumn{3}{c}{Feature Size} \\
    \midrule
    Abbr & Encoder   & LiDAR Range & Voxel Size &  V2V4Real &  OPV2V & V2XSet      \\
		\midrule
		$p_0$         & PointPillars & 140.8, 38.4 & 0.4, 0.4   &  256, 96, 352   &  256, 96, 352  &  256, 96, 352 \\
		$p_1$         & PointPillars & 153.6, 38.4 & 0.6, 0.6  &  256, 64, 256    &  256, 64, 256  &  256, 64, 256  \\
		$s_1$         & SECOND & 140.8, 40 & 0.1, 0.1  &  512, 100, 352   & 256, 100, 352   &  512, 100, 352  \\
		$v_1$         & VoxelNet & 140.8, 40 & 0.8, 0.8 & 512, 100, 352   & 256, 100, 352  &  256, 100, 352  \\
  \bottomrule
  \end{tabular}
  }
    \label{tab:setting}
\end{table}

\subsubsection{Baselines}

Since HEAL \cite{HEAL} and Hetecooper \cite{Hetecooper} require retraining the encoder or designing a new feature fusion module, we only compare the CoDS with plug-and-play adapter-based methods \cite{xu2023bridging,PnPDA,stamp,PolyInter} for a fair comparison. Additionally, we consider a simple baseline HETE, which uses a naive resizer without domain adapters. Note that STAMP \cite{stamp} and PolyInter \cite{PolyInter} were originally designed for open heterogeneous scenarios, where new types of agents with previously unseen models may join the collaboration. Since our focus is on the general heterogeneous setting, where the neighbor agents are fixed but their models are distinct, we adapt and reproduce STAMP and PolyInter accordingly to ensure a fair and consistent comparison.

\begin{itemize}
  \item \textbf{HETE}: It utilizes direct bilinear interpolation and channel slices (or padding) for feature resizing.
  \item \textbf{MPDA} \cite{xu2023bridging}: It employs a learnable feature resizer to resize neighbor features, and a cross-domain transformer to convert the domain of features to the ego domain. 
  \item \textbf{PnPDA} \cite{PnPDA}: It uses a semantic converter to transform neighbor heterogeneous features into the ego domain and a semantic enhancer to strengthen the representation of ego features. Both the converter and enhancer share the same transformer-based architecture.
  \item \textbf{STAMP} \cite{stamp}: It first trains a protocol network, then trains ConvNeXt \cite{liu2022convnet} based local adapters and reverters for feature alignment.
  \item \textbf{PolyInter} \cite{PolyInter}: It projects neighbor features into the ego domain using an interpreter network guided by a general prompt and agent-specific prompts.
  \end{itemize}

\begin{table*}[htbp]
  \large
  \renewcommand\arraystretch{1.3} 
  \centering 
  \caption{Collaborative detection results (AP@0.50$\slash$AP@0.70) in homogeneous scenarios.}
  \resizebox{1\linewidth}{!}{
  \begin{tabular}{c|ccc|ccc|ccc}
    \toprule
Datasets         & \multicolumn{3}{c}{V2V4Real}                                 & \multicolumn{3}{c}{OPV2V}                                    & \multicolumn{3}{c}{V2XSet}                                   \\
  \midrule
  Abbr  & AttFusion         & DiscoNet         & CoBEVT       & AttFusion         & DiscoNet         & CoBEVT         & AttFusion         & DiscoNet         & CoBEVT         \\
  \midrule
  $p_0$             & 73.06$\slash$46.24     & 72.46$\slash$43.33   & 70.70$\slash$41.59    & 93.11$\slash$82.05   & 92.61$\slash$82.29  & 93.52$\slash$78.63 & 88.29$\slash$73.41   & 87.92$\slash$74.09  & 89.05$\slash$74.15  \\
  $p_1$      & 66.63$\slash$35.87   & 67.90$\slash$38.13  & 65.11$\slash$36.14 & 92.81$\slash$80.08  & 92.54$\slash$81.33 &  94.17$\slash$80.10  & 87.74$\slash$68.12  & 88.14$\slash$70.53 &  85.86$\slash$56.30    \\
  $s_1$     & 69.03$\slash$41.87  & 73.25$\slash$50.02 & 55.90$\slash$32.07  & 81.24$\slash$71.21   & 80.93$\slash$71.55 &  90.19$\slash$82.10   & 85.82$\slash$72.19   & 86.39$\slash$73.42 &  81.93$\slash$70.23   \\
  $v_1$      & 67.26$\slash$40.64   & 69.09$\slash$40.53 & 50.98$\slash$31.26  & 76.59$\slash$65.25    & 49.77$\slash$40.96 & 86.31$\slash$73.87  & 80.26$\slash$66.88 & 81.56$\slash$68.07 & 77.71$\slash$64.01    \\
  \bottomrule
  \end{tabular}
  }
    \label{tab:homo}
\end{table*}

\begin{table*}[htbp]
  \Huge
  \renewcommand\arraystretch{1.4} 
  \centering 
  \caption{Detection performance of collaborative detectors. $p_0$, $s_1$ and $v_0$ denotes the collaborative detector in homogeneous scenarios, $p_0$+$p_1$, $p_0$+$s_1$, $p_0$+$v_1$, $s_1$+$p_0$ and $v_1$+$p_0$ denote different heterogeneous scenarios. Results are reported in AP@0.50$\slash$AP@0.70.}
  \resizebox{1\linewidth}{!}{
      \begin{tabular}{c|c|ccc|ccc|ccc}
          \toprule
          \multicolumn{2}{c}{Datasets}      & \multicolumn{3}{c}{V2V4Real}                                      & \multicolumn{3}{c}{OPV2V}                                         & \multicolumn{3}{c}{V2XSet}                                        \\
          \midrule
          Encoder               & Method   & AttFusion       & DiscoNet        & CoBEVT         & AttFusion       & DiscoNet        & CoBEVT        & AttFusion       & DiscoNet        & CoBEVT       \\
          \midrule
          $p_0$                     &     -    & 73.06$\slash$46.24  & 72.46$\slash$43.33 & 70.70$\slash$41.59  & 93.11$\slash$82.05  & 92.61$\slash$82.29 & 93.52$\slash$78.63  & 88.29$\slash$73.41  & 87.92$\slash$74.09 &  89.05$\slash$74.15       \\
          \midrule
          \multirow{6}{*}{$p_0$+$p_1$} & HETE    & 55.41$\slash$36.59  & 50.96$\slash$31.19 & 33.19$\slash$5.85 & 83.86$\slash$70.07  & 86.64$\slash$76.28 & 66.63$\slash$42.72 & 80.56$\slash$66.42 & 81.15$\slash$65.43 &  79.06$\slash$50.78       \\
                                 & MPDA     & 59.34$\slash$36.92     & 58.45$\slash$35.51  & 60.34$\slash$35.31  & 88.22$\slash$77.55 & 87.69$\slash$77.31   & 79.13$\slash$69.99    & 83.04$\slash$68.96    & 81.48$\slash$67.11 & 85.46$\slash$70.87    \\
                      & PnPDA     & 63.99$\slash$32.63     & 60.08$\slash$\textbf{38.77}  &  \textbf{63.53}$\slash$35.61 & 87.31$\slash$74.18     & 76.65$\slash$55.98  &  66.24$\slash$55.71 & 82.21$\slash$67.27     & 69.06$\slash$50.04  & 74.05$\slash$56.20   \\
                      & STAMP & \textbf{64.08}$\slash$\textbf{40.93} & \textbf{63.59}$\slash$37.17 & 60.07$\slash$35.24 & 85.42$\slash$61.36 & 80.78$\slash$64.63 & 86.53$\slash$75.09 & 82.42$\slash$63.59 & 81.89$\slash$58.70 & 85.67$\slash$71.76   \\
                      & PolyInter & 56.35$\slash$37.79 & 59.32$\slash$37.81 & 55.57$\slash$27.75 & 88.07$\slash$77.53 & 84.63$\slash$74.49 & \textbf{88.66}$\slash$\textbf{79.07} & 82.83$\slash$68.94 & 81.70$\slash$\textbf{69.14} & 69.16$\slash$58.19   \\
                                 & CoDS (Ours)    & 60.85$\slash$37.21 & 60.99$\slash$37.89 & 61.62$\slash$\textbf{40.30} & \textbf{88.81}$\slash$\textbf{77.55}   & \textbf{87.97}$\slash$\textbf{77.49} & 86.84$\slash$74.48 & \textbf{84.03}$\slash$\textbf{69.39} & \textbf{82.71}$\slash$67.11 & \textbf{86.24}$\slash$\textbf{72.37}\\
          \midrule
          \multirow{6}{*}{$p_0$+$s_1$} & HETE    & 55.45$\slash$36.58   & 41.21$\slash$28.84 & 12.10$\slash$6.51   & 87.96$\slash$76.30 & 86.44$\slash$76.03  & 62.86$\slash$33.63 & 82.87$\slash$68.21  & 81.31$\slash$67.81  &  76.32$\slash$45.46   \\
                                 & MPDA     & 59.61$\slash$37.40      & 67.24$\slash$40.05 &  65.94$\slash$35.10   & 91.99$\slash$79.57   & 91.18$\slash$77.72 & 91.47$\slash$77.87 & 87.42$\slash$73.11   & 87.04$\slash$71.32&  83.54$\slash$65.83   \\
                                 & PnPDA     & 68.38$\slash$39.60     & 68.90$\slash$44.98  &  64.82$\slash$37.24 & 83.34$\slash$71.99    & 75.72$\slash$56.36  &  \textbf{92.85}$\slash$79.48 & 77.33$\slash$62.06     & 76.03$\slash$64.39  &  59.96$\slash$47.64  \\
                                 & STAMP & \textbf{68.53}$\slash$43.30 & 70.59$\slash$42.05 & 65.94$\slash$38.43  & 79.54$\slash$62.61 & 90.59$\slash$72.51 & 86.01$\slash$75.86 & 87.31$\slash$60.06 & 87.99$\slash$68.16 & 88.04$\slash$73.96    \\
                                 & PolyInter & 65.17$\slash$36.25 &  66.83$\slash$42.31 & 65.17$\slash$36.25  & 92.50$\slash$\textbf{81.47} & 90.53$\slash$78.18 & 92.47$\slash$\textbf{82.72}& 89.09$\slash$\textbf{75.61} & 88.91$\slash$76.51 & 82.15$\slash$68.50    \\
                                 & CoDS (Ours)    & 65.37$\slash$\textbf{44.21} & \textbf{71.27}$\slash$\textbf{46.52} & \textbf{69.00}$\slash$\textbf{38.58} & \textbf{92.83}$\slash$79.86 & \textbf{92.11}$\slash$\textbf{81.76} & 90.13$\slash$77.49 & \textbf{89.14}$\slash$74.61 & \textbf{90.12}$\slash$\textbf{78.33} & \textbf{88.63}$\slash$\textbf{76.06}\\
          \midrule
          \multirow{6}{*}{$p_0$+$v_1$} & HETE    & 55.42$\slash$36.60   & 33.93$\slash$24.13  & 46.33$\slash$16.71 & 87.95$\slash$76.30    & 86.89$\slash$76.37  & 90.09$\slash$78.45 & 82.87$\slash$68.21   & 79.38$\slash$66.59  & 80.12$\slash$52.34 \\
                                 & MPDA     & 59.39$\slash$38.08   & 65.08$\slash$37.49  & 67.96$\slash$37.57 & 89.96$\slash$77.68   & 89.96$\slash$77.68   & 90.18$\slash$\textbf{80.59}  & 85.78$\slash$70.47  & 86.88$\slash$73.33 &   50.83$\slash$38.89       \\
                                 & PnPDA     & 68.26$\slash$43.38     & 65.58$\slash$35.97  &  67.14$\slash$38.03  & 74.07$\slash$64.44     & 77.11$\slash$61.95  & 79.45$\slash$65.13  & 55.09$\slash$45.43 & 78.43$\slash$61.56  &  49.52$\slash$42.44   \\
                                 & STAMP & \textbf{69.92}$\slash$\textbf{45.09} & 69.25$\slash$42.55 & 67.28$\slash$38.10 & 87.26$\slash$56.37 & 90.99$\slash$70.79 & \textbf{91.51}$\slash$76.91 & 86.57$\slash$63.08 & 87.15$\slash$67.29 & 87.98$\slash$73.76    \\
                      & PolyInter & 68.53$\slash$43.30 & 66.67$\slash$39.08 & 61.29$\slash$32.32  & 91.97$\slash$79.50 & 89.25$\slash$73.70 & 90.14$\slash$79.84 & 88.20$\slash$73.29 & \textbf{87.25}$\slash$\textbf{74.70} & 78.47$\slash$65.47    \\
                                 & CoDS (Ours)     & 66.04$\slash$44.81 & \textbf{69.78}$\slash$\textbf{43.41} & \textbf{68.38}$\slash$\textbf{39.87} & \textbf{93.63}$\slash$\textbf{82.45} & \textbf{93.35}$\slash$\textbf{84.07} & 90.24$\slash$80.00& \textbf{88.69}$\slash$\textbf{76.11} & 87.16$\slash$73.73 & \textbf{88.32}$\slash$\textbf{73.81}\\
      \midrule
      $s_1$                     &     -    & 69.03$\slash$41.87  & 73.25$\slash$50.02 & 55.90$\slash$32.07  & 81.24$\slash$71.21   & 80.93$\slash$71.55 &  90.19$\slash$82.10   & 85.82$\slash$72.19   & 86.39$\slash$73.42 &  81.93$\slash$70.23       \\
              \midrule
              \multirow{6}{*}{$s_1$+$p_0$} & HETE   & 54.17$\slash$34.77 & 55.36$\slash$40.73 & 51.77$\slash$27.97 &  81.49$\slash$71.39 & 74.68$\slash$65.85  & 88.34$\slash$78.51 & 81.07$\slash$68.74  & 73.91$\slash$64.53 & 83.08$\slash$66.38   \\
                                     & MPDA     & 63.82$\slash$36.32 & \textbf{68.58}$\slash$42.85  & 67.46$\slash$34.74 & 89.73$\slash$75.21  & 87.05$\slash$70.64  &  89.56$\slash$81.04& 82.01$\slash$70.24  & 80.34$\slash$69.35 & 86.03$\slash$73.13 \\
                                     & PnPDA     & 60.73$\slash$37.92 & 60.21$\slash$38.10  & 62.58$\slash$31.12 & 61.18$\slash$51.43  & 79.97$\slash$48.95  & 92.69$\slash$79.10 & \textbf{87.35}$\slash$72.63  & 78.41$\slash$68.53 & 86.34$\slash$72.59  \\
                                     & STAMP & 62.18$\slash$40.89 & 67.77$\slash$37.41 & 66.97$\slash$33.92  & 83.55$\slash$67.35 & 82.51$\slash$62.19 & 93.19$\slash$83.29 & 80.02$\slash$70.32 & 60.55$\slash$41.87 & 38.60$\slash$34.12   \\
                      & PolyInter & 63.35$\slash$41.97 & 65.36$\slash$\textbf{45.99} & 62.28$\slash$34.50 & \textbf{90.85}$\slash$77.08 & 86.97$\slash$71.05 & 66.63$\slash$60.26 & 86.42$\slash$\textbf{73.19} & 73.92$\slash$66.06 & 80.44$\slash$69.41  \\
                                     & CoDS (Ours)    & \textbf{69.79}$\slash$\textbf{46.58} & 66.07$\slash$45.34  & \textbf{67.47}$\slash$\textbf{36.49} &  86.76$\slash$\textbf{77.44} & \textbf{90.59}$\slash$\textbf{78.34}  & \textbf{93.57}$\slash$\textbf{84.01} &83.40$\slash$72.66  & \textbf{85.91}$\slash$\textbf{71.46} & \textbf{86.92}$\slash$\textbf{73.52} \\
              \midrule
              $v_1$                     &     -    & 67.26$\slash$40.64   & 69.09$\slash$40.53 & 50.98$\slash$31.26  & 76.59$\slash$65.25  & 49.77$\slash$40.96 & 86.31$\slash$73.87  & 80.26$\slash$66.88 & 81.56$\slash$68.07 & 77.71$\slash$64.01     \\
              \midrule
              \multirow{6}{*}{$v_1$+$p_0$} & HETE   & 51.23$\slash$33.48 & 52.45$\slash$32.47 & 40.98$\slash$18.70 & 78.19$\slash$65.90  & 77.45$\slash$63.82 & 82.86$\slash$65.11 & 74.34$\slash$62.95  &76.15$\slash$62.38 & 52.19$\slash$40.39   \\
                                     & MPDA     & 65.21$\slash$39.51 & 64.27$\slash$36.45  & 62.07$\slash$34.95 & 88.59$\slash$70.59  & 79.85$\slash$66.90  &  90.96$\slash$77.78 & 83.92$\slash$67.84  & 78.25$\slash$66.66 &  70.31$\slash$49.14     \\
                                     & PnPDA     & 57.69$\slash$34.23 & 63.07$\slash$32.60  & 54.13$\slash$29.72 & 62.43$\slash$50.76  & 86.15$\slash$64.21  & 85.39$\slash$73.98 & 86.42$\slash$70.72  & 82.18$\slash$69.59 &  84.71$\slash$68.52   \\
                                     & STAMP & 66.78$\slash$39.23 & 65.75$\slash$27.94 & \textbf{62.99}$\slash$\textbf{37.66}  & 84.02$\slash$52.75 & 82.27$\slash$60.98 & 91.68$\slash$79.20 & 83.71$\slash$64.88 & 77.97$\slash$57.55 & 85.01$\slash$68.57    \\
                      & PolyInter & 64.30$\slash$40.73 & 63.83$\slash$38.20 & 58.72$\slash$33.35  & \textbf{91.45}$\slash$\textbf{79.51} & 87.38$\slash$68.85 & 90.51$\slash$76.73 & \textbf{86.98}$\slash$\textbf{73.92} & \textbf{86.36}$\slash$\textbf{73.19} & 85.02$\slash$69.53   \\
                                     & CoDS (Ours)    & \textbf{67.00}$\slash$\textbf{42.60} & \textbf{67.41}$\slash$\textbf{41.26} & 62.13$\slash$35.87 & 89.13$\slash$74.41 & \textbf{92.06}$\slash$\textbf{77.34} & \textbf{91.97}$\slash$\textbf{79.49} & 84.70$\slash$65.83 & 85.26$\slash$70.27 & \textbf{85.52}$\slash$\textbf{69.71}\\
        \bottomrule
          \end{tabular}
  }
  
  \label{tab:hetero}
\end{table*}

\subsection{Quantitative Evaluation}
\subsubsection{Accuracy Comparison}
Table~\ref{tab:hetero} shows the detection accuracy in different datasets, collaborative detectors and heterogeneous scenarios, where $p_0$ denotes collaborative detectors in homogeneous scenarios. In contrast, $p_0$+$p_1$ indicates the ego is equipped with the $p_0$ encoder while neighbor agents are equipped with the $p_1$ encoder.

In V2V4Real, when collaborative detectors are AttFusion or DiscoNet, the accuracy of HETE decreases by approximately 10 in AP@0.70 compared to homogeneous scenarios. For CoBEVT, this drop increases to about 20 in AP@0.70. Adapter-based methods such as MPDA, PnPDA, STAMP, PolyInter and our CoDS enhance the accuracy of collaborative detectors in heterogeneous scenarios, with CoDS demonstrating the most consistent improvements. In particular, combining DiscoNet with CoDS yields an average gain of 20.32 in AP@0.50 and 11.39 in AP@0.70 compared to HETE. In addition, CoDS enables collaborative detectors in heterogeneous scenarios ($s_0$+$p_0$ and $v_0$+$p_1$) to achieve higher AP@0.70 than those in homogeneous scenarios ($s_1$ and $v_0$).

In OPV2V and V2XSet, the accuracy of HETE in heterogeneous scenarios decreases by approximately 10 to 30 AP@0.70 compared to homogeneous scenarios. However, in OPV2V, when the collaborative detectors are AttFusion and DiscoNet, HETE achieves higher accuracy in the $v_1$+$p_0$ setting compared to $v_1$. This is because the feature discrepancies in these scenarios are minimal, allowing heterogeneous features to still achieve effective complementarity. In heterogeneous scenarios, MPDA, PnPDA, STAMP and PolyInter may exhibit unstable performance, occasionally performing worse than HETE. In contrast, our CoDS consistently outperforms previous methods across most settings, highlighting the effectiveness of domain separation in addressing feature discrepancies.

\begin{table*}[htbp]
  \small
  \renewcommand\arraystretch{1.23} 
  \centering 
  \caption{Inference speed (FPS) under different collaboration numbers.}
  \resizebox{0.95\linewidth}{!}{
    \begin{tabular}{c|c|c|cccc|cccc}
      \toprule
    \multicolumn{2}{c}{Datasets}       & \multicolumn{1}{c}{V2V4Real}  & \multicolumn{4}{c}{OPV2V}    & \multicolumn{4}{c}{V2XSet}                                                                                                  \\
      \midrule
      \multicolumn{2}{c}{Agent Numbers} & 2 Agents                      & 2 Agents                      & 3 Agents                      & 4 Agents                      & 5 Agents                      & 2 Agents                      & 3 Agents                      & 4 Agents                      & 5 Agents                      \\
      \midrule
      \multirow{5}{*}{$p_0$+$p_1$}   & 1. MPDA   & 28.74           & 30.69                         & 21.64                         & 17.20    & 14.78   & 31.19                         & 18.52                         & 15.77                         & 13.59                        \\
                                & 2. PnPDA   &  29.58 & 34.29 & 27.21 & 24.35 & 22.15 & 40.36 & 31.10 & 25.81 & 23.53    \\
                                & 3. STAMP  & 39.35 & 39.38 & 32.60 & 28.94 & 26.01  &  40.20 & 29.90 & 26.94 & 24.36 \\
                                & 4. PolyInter  & 39.59 & 40.87 & 28.96 & 24.28 & 20.86  &  41.19 & 25.56 & 21.70 & 18.97 \\
                               & 5. CoDS (Ours)  & \textbf{46.99} & \textbf{47.58} & \textbf{39.67} & \textbf{35.01} & \textbf{33.06} & \textbf{48.99} & \textbf{37.87} & \textbf{34.43} & \textbf{31.97}  \\
                               \midrule
  \multirow{5}{*}{$p_0$+$s_1$}& 1. MPDA   & 28.88   & 27.33                         & 18.56                         & 14.86                         & 12.72   & 29.59                         & 17.23                         & 14.66                         & 12.75         \\
                              & 2. PnPDA   &  33.76 & 29.63 & 23.36 & 20.65 & 18.66 & 40.00 & 33.38 & 30.26 & 27.42   \\
                              & 3. STAMP  & 41.54 & 36.19 & 29.05 & 25.48 & 22.70 & 40.69 & 31.24 & 28.39 & 26.03 \\
                                & 4. PolyInter  & 41.50 & 36.39  &  26.05 & 21.75 & 18.72 &  40.75 & 26.24 & 22.70 & 20.00 \\
                               & 5. CoDS (Ours)  & \textbf{50.19} & \textbf{43.56} & \textbf{35.47} & \textbf{31.17} & \textbf{28.16} & \textbf{48.82} & \textbf{38.17} & \textbf{35.82} & \textbf{33.44}\\
                               \midrule
   \multirow{5}{*}{$p_0$+$v_1$}    & 1. MPDA   & 29.03       & 29.76               & 20.44                         & 16.87                         & 14.20   & 29.94                         & 17.40                         & 14.75                         & 12.87                           \\
                            & 2. PnPDA   &   33.58 & 35.32 & 28.64 & 25.58 & 21.52 & 42.65 & 33.04 & 30.94 & 28.22 \\
                            & 3. STAMP  & 40.78 & 40.91  & 34.20 & 30.77 & 28.06  & 41.39 & 31.61 & 28.77 & 26.45 \\
                                & 4. PolyInter  & \textbf{41.49} & 41.35  &  30.20 & 25.51 & 22.18  &  41.47 & 26.63 & 23.05 & 20.35 \\
                               & 5. CoDS (Ours)   & 39.55 & \textbf{49.02} & \textbf{40.56} & \textbf{39.39} & \textbf{35.86} & \textbf{49.59} & \textbf{38.43} & \textbf{37.77} & \textbf{35.06}  \\
                               \midrule
\multirow{5}{*}{$s_1$+$p_0$}    & 1. MPDA   &  22.76 & 22.66 & 16.30  & 13.47 & 11.35 & 22.87 & 13.48 & 11.76 & 10.12  \\
                            & 2. PnPDA   &  21.93 & 30.71 & 25.58 & 21.24 & 17.76 & 29.31 & 24.01 & 20.52 & 16.56  \\
                            & 3. STAMP  & 16.72 & 29.61  &  24.45 & 21.64 & 19.16  &  28.67  &  24.55 & 22.34 & 19.27  \\
                                & 4. PolyInter & 18.61 & 30.09 &  22.32 & 18.93 & 16.11 & 29.87 & 22.32 & 17.23 & 15.10 \\
                               & 5. CoDS (Ours)   & \textbf{24.03} & \textbf{33.43} & \textbf{28.49} & \textbf{25.74} & \textbf{22.96} & \textbf{34.04} & \textbf{28.78} & \textbf{25.09} & \textbf{22.04} \\
                               \midrule
\multirow{5}{*}{$v_1$+$p_0$}    & 1. MPDA   &  22.64          & 23.64          & 16.53          & 13.33          & 11.42          & 22.71          & 13.84          & 11.81          & 10.30     \\
                               & 2. PnPDA   &  21.98 & 28.23 & 23.82 & 22.40 & 19.68 & 29.16 & 23.36 & 20.62 & 18.82 \\
                               & 3. STAMP  & 13.45 & 27.67 & 23.37 & 21.03 & 18.67  &  27.77 & 21.51 & 19.59 & 17.34 \\
                                & 4. PolyInter  & 14.82 & 28.77 & 21.56 & 18.61 & 16.01  & 28.63 & 19.20 & 16.82 & 14.50 \\
                                  & 5. CoDS (Ours)   & \textbf{24.21} & \textbf{31.59} & \textbf{28.46} & \textbf{26.00} & \textbf{22.87} & \textbf{34.24} & \textbf{26.35} & \textbf{24.75} & \textbf{20.27} \\
                               \bottomrule
  \end{tabular}
}
\label{tab:fps}
\end{table*}

\begin{figure*}[htbp]
  \centerline{\includegraphics[width=0.8\linewidth]{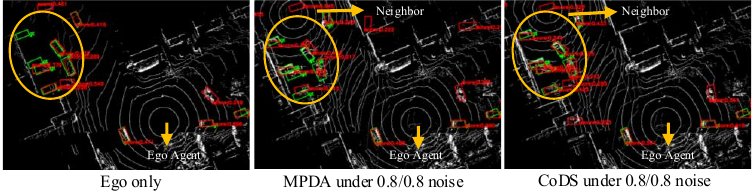}}
    \caption{\textbf{Impact of pose error in heterogeneous scenarios.} Large localization noise causes severe bounding box misalignments in collaborative perception relative to the ego-only baseline.}
    \label{fig:pose}
  \end{figure*}

\begin{figure*}[htbp]
    \centerline{\includegraphics[width=\linewidth]{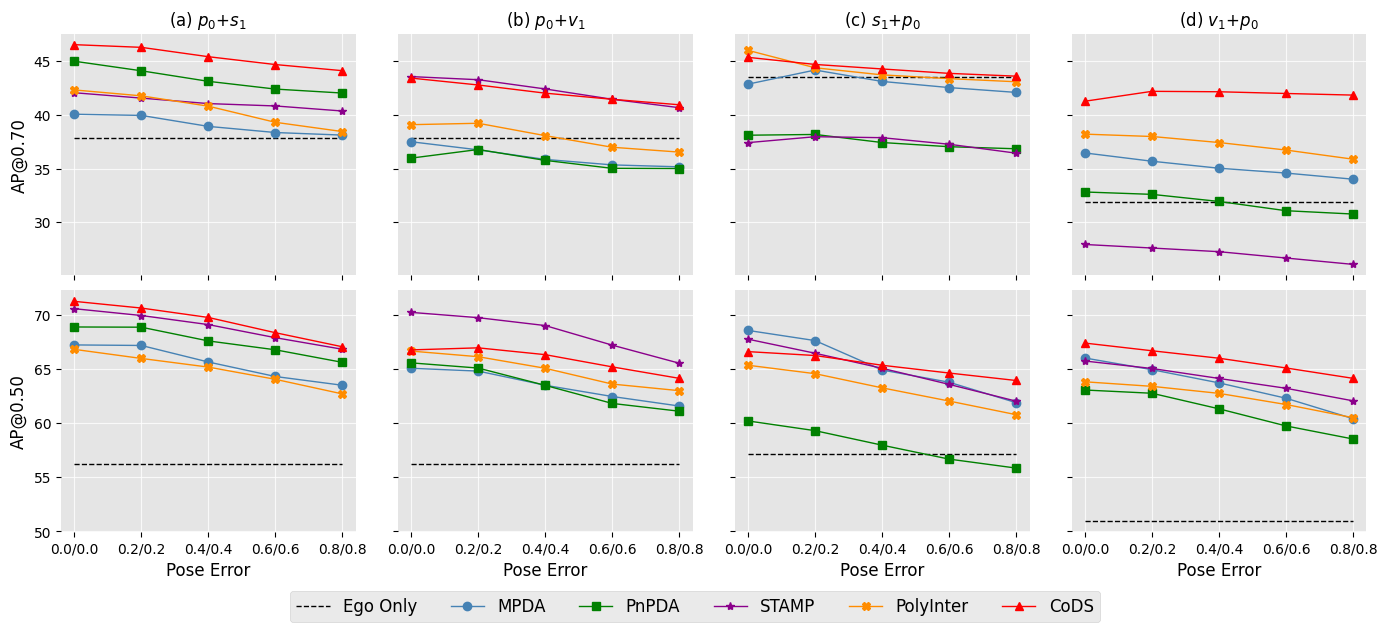}}
      \caption{\textbf{Robust Experiment to pose error on V2V4Real}. Pose noise is set to $\mathcal{N}(0,\sigma_p^2)$ on the x, y position and $\mathcal{N}(0,\sigma_r^2)$ on the yaw angle. CoDS achieves state-of-the-art performance under various noisy conditions, consistently surpassing individual perception (ego only).}
      \label{loc_error}
\end{figure*}

\subsubsection{Efficiency Comparison} We evaluate the inference efficiency of adapter-based methods (MPDA, PnPDA, STAMP, PolyInter and CoDS) in heterogeneous scenarios and examine the impact of the number of agents on inference efficiency. Specifically, we use 2 agents in V2V4Real and 2 to 5 agents in OPV2V and V2XSet. 

The results in Table~\ref{tab:fps} indicate that with a small number of collaborators, the collaborative detectors with CoDS exhibit significantly higher inference speeds than previous methods. Specifically, when there is only one neighbor agent, the CoDS outperforms MPDA and PnPDA by over $30\%$ in FPS.

As the number of agents increases, the FPS of collaborative detectors decreases due to the additional computational requirements for aligning and fusing more features. Despite this, CoDS maintains a significant inference advantage over previous methods. Specifically, when the maximum number of agents reaches five, CoDS achieves an FPS improvement of over $100\%$ compared to MPDA and more than $20\%$ compared to PnPDA, STAMP and PolyInter. These improvements are largely attributed to the fully convolutional architecture of CoDS, which ensures relatively low inference costs.

\begin{table}[htbp]
  \Huge
	\centering 
  \renewcommand\arraystretch{1.2} 
	\caption{Parameter sizes (M) of different adapter-based models.}
	\resizebox{1\linewidth}{!}{
		\begin{tabular}{cccccc}
			\toprule
			Methods & MPDA & PnPDA & STAMP & PolyInter & CoDS (Ours)  \\
			\midrule
			Params (M) & 6.12 &  \textbf{3.29}  & 4.81 & 46.22 & 3.67 \\
			\bottomrule
			\end{tabular}
	}
	\label{tab:param}
  \vspace{-0.5em}
\end{table}

We also report the parameter sizes of different adapter-based models in Table~\ref{tab:param}. The results show that our CoDS requires only 3.67M parameters, significantly smaller than PolyInter (46.22M) and competitive with other efficient adapter-based methods, such as PnPDA (3.29M).

\subsubsection{Localization Error Robustness}
To effectively share valid information, multiple agents require accurate poses to synchronize their individual data within a consistent spatial coordinate system. However, the 6-DoF poses estimated by each agent's localization module are not always perfect in practice, leading to relative pose inaccuracies.
As shown in Fig.~\ref{fig:pose}, adapter-based methods (MPDA and CoDS) detect more objects than the ego-only (no fusion) baseline but suffer from severe bounding box misalignments, with some predictions even deviating farther from the ground-truth than the ego vehicle alone.
Therefore, we further evaluate the performance of CoDS and other adapter-based methods in heterogeneous scenarios with localization errors, as illustrated in Fig.~\ref{loc_error}. To simulate localization errors, Gaussian noise $\mathcal{N}(0,\sigma_p^2)$ is added to the 2D center coordinates $x$ and $y$, and $\mathcal{N}(0,\sigma_r^2)$ is added to the yaw angle $\theta$, where $x$, $y$ and $\theta$ represent the accurate global pose parameters.

When there is no localization noise, all methods achieve a high AP@0.50. As localization errors increase, the performance of all methods declines, occasionally falling below the accuracy of individual perception (ego only). This will affect the safety of autonomous driving. However, CoDS consistently outperforms the other methods and maintains higher performance than individual perception. This is because, under the guidance of DAMI loss, CoDS is still able to capture task-related information even in the presence of localization errors.

\begin{figure*}[htbp]
  \centering
  \begin{minipage}{0.01\textwidth}
    \centering
    \makebox[1\linewidth]{\scriptsize\rotatebox{90}{Before}}
  \end{minipage}
  \begin{minipage}{0.323\linewidth}
    \centerline{\includegraphics[width=\linewidth]{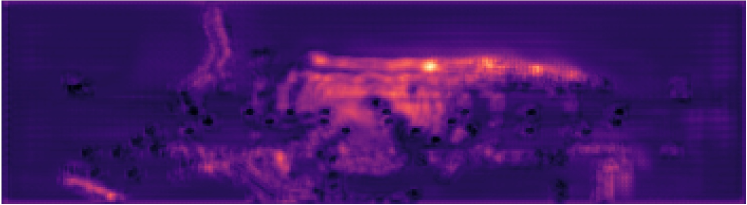}}
  \end{minipage}
  \begin{minipage}{0.323\linewidth}
    \centerline{\includegraphics[width=\linewidth]{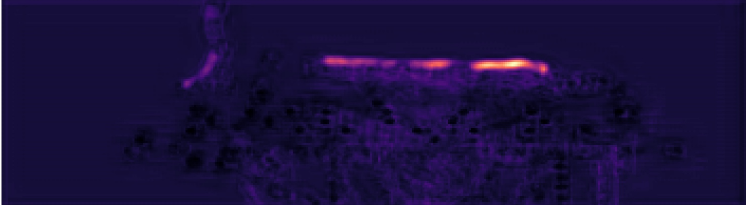}}
  \end{minipage}
  \begin{minipage}{0.323\linewidth}
    \centerline{\includegraphics[width=\linewidth]{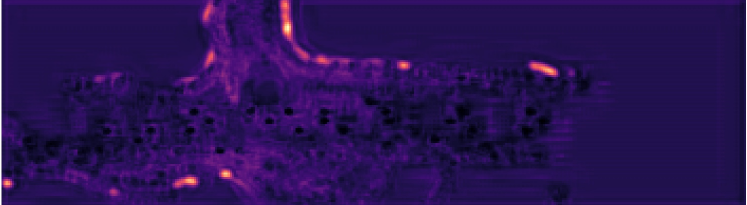}}
  \end{minipage}
  \vspace{0.4\baselineskip} 
  \\
  \begin{minipage}{0.01\textwidth}
    \centering
    \makebox[1\linewidth]{\scriptsize\rotatebox{90}{After}}
  \end{minipage}
  \begin{minipage}{0.323\linewidth}
    \centerline{\includegraphics[width=\linewidth]{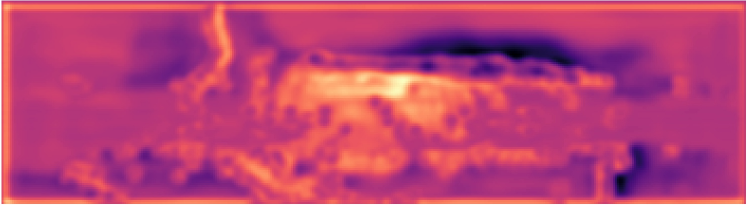}}
    \small\centerline{(a) Ego vehicle with $p_0$}
  \end{minipage}
  \begin{minipage}{0.323\linewidth}
    \centerline{\includegraphics[width=\linewidth]{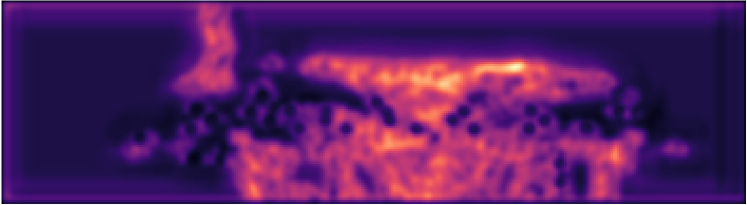}}
    \small\centerline{(b) Neighbor Agent 1 with $s_1$ }
  \end{minipage}
  \begin{minipage}{0.323\linewidth}
    \centerline{\includegraphics[width=\linewidth]{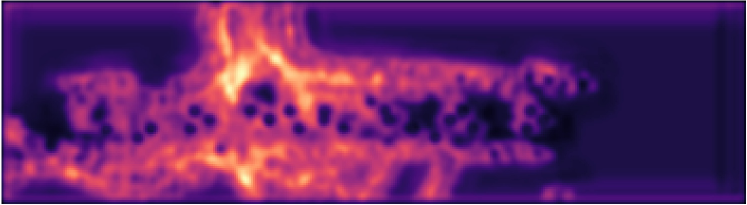}}
    \small\centerline{(c) Neighbor Agent 2 with $s_1$}
  \end{minipage}
\caption{\textbf{Visualization of intermediate features before and after alignment.} Ego and neighbor agents use PointPillars and SECOND encoders, respectively. After processing by CoDS, the heterogeneous features exhibit similar semantic characteristics.}
\vspace{-0.5em}
\label{feat_vis}
\end{figure*}

\begin{figure*}[htbp]
  \centering
  \begin{minipage}{0.01\textwidth}
    \centering
    \makebox[1\linewidth]{\scriptsize\rotatebox{90}{HETE}}
  \end{minipage}
  \begin{minipage}{0.323\linewidth}
    \centerline{\includegraphics[width=\linewidth]{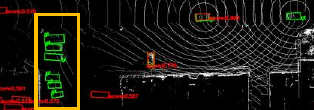}}
  \end{minipage}
  \begin{minipage}{0.323\linewidth}
    \centerline{\includegraphics[width=\linewidth]{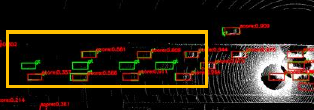}}
  \end{minipage}
  \begin{minipage}{0.323\linewidth}
    \centerline{\includegraphics[width=\linewidth]{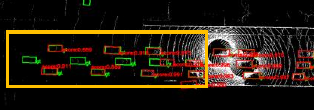}}
  \end{minipage}
  \vspace{0.4\baselineskip} 
  \\
  \begin{minipage}{0.01\textwidth}
    \centering
    \makebox[1\linewidth]{\scriptsize\rotatebox{90}{MPDA}}
  \end{minipage}
  \begin{minipage}{0.323\linewidth}
    \centerline{\includegraphics[width=\linewidth]{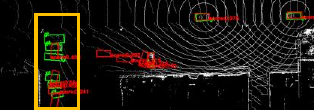}}
  \end{minipage}
  \begin{minipage}{0.323\linewidth}
    \centerline{\includegraphics[width=\linewidth]{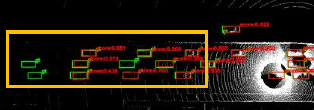}}
  \end{minipage}
  \begin{minipage}{0.323\linewidth}
    \centerline{\includegraphics[width=\linewidth]{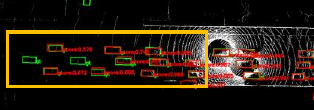}}
  \end{minipage}
  \vspace{0.4\baselineskip} 
  \\
  \begin{minipage}{0.01\textwidth}
    \centering
    \makebox[1\linewidth]{\scriptsize\rotatebox{90}{PnPDA}}
  \end{minipage}
  \begin{minipage}{0.323\linewidth}
    \centerline{\includegraphics[width=\linewidth]{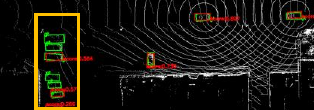}}
  \end{minipage}
  \begin{minipage}{0.323\linewidth}
    \centerline{\includegraphics[width=\linewidth]{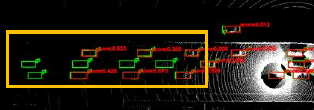}}
  \end{minipage}
  \begin{minipage}{0.323\linewidth}
    \centerline{\includegraphics[width=\linewidth]{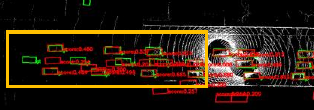}}
  \end{minipage}
  \vspace{0.4\baselineskip} 
  \\
  \begin{minipage}{0.01\textwidth}
    \centering
    \makebox[1\linewidth]{\scriptsize\rotatebox{90}{STAMP}}
  \end{minipage}
  \begin{minipage}{0.323\linewidth}
    \centerline{\includegraphics[width=\linewidth]{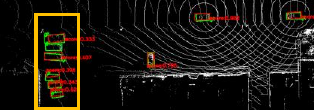}}
  \end{minipage}
  \begin{minipage}{0.323\linewidth}
    \centerline{\includegraphics[width=\linewidth]{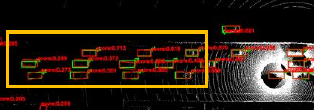}}
  \end{minipage}
  \begin{minipage}{0.323\linewidth}
    \centerline{\includegraphics[width=\linewidth]{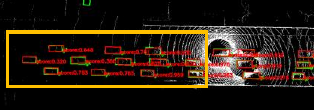}}
  \end{minipage}
  \vspace{0.4\baselineskip} 
  \\
  \begin{minipage}{0.01\textwidth}
    \centering
    \makebox[1\linewidth]{\scriptsize\rotatebox{90}{PolyInter}}
  \end{minipage}
  \begin{minipage}{0.323\linewidth}
    \centerline{\includegraphics[width=\linewidth]{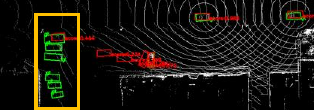}}
  \end{minipage}
  \begin{minipage}{0.323\linewidth}
    \centerline{\includegraphics[width=\linewidth]{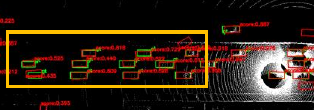}}
  \end{minipage}
  \begin{minipage}{0.323\linewidth}
    \centerline{\includegraphics[width=\linewidth]{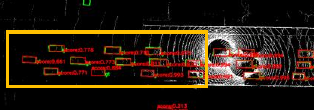}}
  \end{minipage}
  \vspace{0.4\baselineskip} 
  \\
  \begin{minipage}{0.01\textwidth}
    \centering
    \makebox[1\linewidth]{\scriptsize\rotatebox{90}{CoDS}}
  \end{minipage}
  \begin{minipage}{0.323\linewidth}
    \centerline{\includegraphics[width=\linewidth]{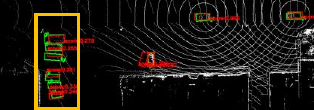}}
    \small\centerline{(a) V2V4Real}
  \end{minipage}
  \begin{minipage}{0.323\linewidth}
    \centerline{\includegraphics[width=\linewidth]{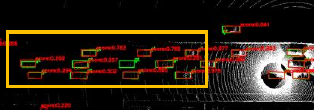}}
    \small\centerline{(b) OPV2V}
  \end{minipage}
  \begin{minipage}{0.323\linewidth}
    \centerline{\includegraphics[width=\linewidth]{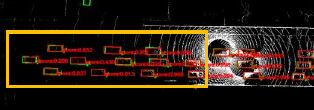}}
    \small\centerline{(c) V2XSet}
  \end{minipage}
\caption{\textbf{Visualization of detection results using different collaborative methods.} The \textcolor{green}{green} and \textcolor{red}{red} boxes represent the ground truth and the detection results by different collaboration methods, respectively. The proposed CoDS achieves the most detection performance across various datasets.}
\vspace{-0.5em}
\label{det_vis}
\end{figure*}

\subsection{Qualitative Evaluation}

\subsubsection{Visualization of Feature Maps} 
Fig.~\ref{feat_vis} illustrates the feature maps before and after alignment by CoDS. Before alignment, there are substantial differences in the original semantics of ego and neighbor features. For PointPillars, the foreground regions on the feature map exhibit relatively higher values, whereas for SECOND, the foreground regions on the feature map exhibit relatively lower values.

However, after processing with CoDS, the patterns of ego and neighbor features become noticeably more similar, exhibiting consistent color characteristics. All aligned features emphasize the object regions, demonstrating that CoDS effectively removes encoder-specific information while capturing task-related information. This highlights the effectiveness of CoDS in addressing distribution discrepancies.

\subsubsection{Visualization of Detection Results} 
We visualize the detection results of different methods across three datasets, where the ego and neighbor agents utilize PointPillars and SECOND encoders, respectively.
As shown in Fig.~\ref{det_vis}, when HETE is directly applied for collaboration, the detector misses numerous objects and even produces significant false detections in V2V4Real. This highlights the negative impact of feature discrepancies on intermediate collaboration.

Collaborative detectors employing various adapter-based methods demonstrate improved object detection. However, due to ambiguities in feature fusion caused by distribution differences, the detectors using MPDA and PnPDA fail to identify certain regions, while STAMP and PolyInter tend to produce false detections. In contrast, our CoDS method significantly reduces these missed detections, demonstrating its effectiveness in addressing distribution discrepancies.

\begin{table}[htbp]
  \large
  \renewcommand\arraystretch{1.4} 
  \centering 
  \caption{Ablation study of the components on V2V4Real.}
  \resizebox{0.85\linewidth}{!}{
      \begin{tabular}{cccccc}
          \toprule
          Encoder   & $\boldsymbol{F}_{\text{LSCR}}$ & $\boldsymbol{F}_{\text{DADS}}$ &  $\mathcal{L}_{\text{DAMI}}$  & AP@0.50 & AP@0.70 \\
          \midrule
          \multirow{5}*{$p_0$+$p_1$}      &   - & - & -  & 50.96  & 31.19  \\
                &  \checkmark & - & -    & 59.96  & 36.99  \\
                &  \checkmark & \checkmark & -     & 60.38  & 34.51  \\
                &  \checkmark & -  & \checkmark    & 60.83  & 37.89  \\
                &  \checkmark & \checkmark  & \checkmark  & \textbf{60.99}  & \textbf{37.89}  \\
          \midrule
          \multirow{5}*{$p_0$+$s_1$}       &   - & - & -    & 41.21  & 28.84  \\
                &  \checkmark & - & -    & 69.41  & 39.63  \\
                &  \checkmark & \checkmark & -     & 68.32    & 37.95   \\
                &  \checkmark & -  & \checkmark    & 68.42  & 37.63  \\
                &  \checkmark & \checkmark  & \checkmark   & \textbf{71.27}  & \textbf{46.52}  \\
          \midrule
          \multirow{5}*{$p_0$+$v_1$}       &   - & - & -     & 33.93  & 24.13  \\
                &  \checkmark & - & -     & 69.57  & 39.61 \\
                &  \checkmark & \checkmark & -    & 68.24  & 42.50  \\
                &  \checkmark & -  & \checkmark    & 69.23  & 39.68  \\
                &  \checkmark & \checkmark  & \checkmark  & \textbf{69.78}  & \textbf{43.41}  \\
          \bottomrule
          \end{tabular}
  }
  
  \label{tab:Ablation_Module}
  \vspace{-1em}
\end{table}

\subsection{Ablation Studies} \label{Ablation}

\subsubsection{Contribution of Components} 
We conduct ablation studies to evaluate the effectiveness of the proposed LSCR module $\boldsymbol{F}_{\text{LSCR}}$, DADS module $\boldsymbol{F}_{\text{DADS}}$ and DAMI loss $\mathcal{L}_{\text{DAMI}}$. We take HETE as a baseline and gradually incorporate each component. As shown in Table~\ref{tab:Ablation_Module}, using $\boldsymbol{F}_{\text{LSCR}}$ effectively addresses the dimension discrepancy issue and improves the AP@0.70 over the baseline over 18$\%$. However, when $\boldsymbol{F}_{\text{LSCR}}$ is combined with $\boldsymbol{F}_{\text{DADS}}$ without $\mathcal{L}_{\text{DAMI}}$, the performance decreases due to a lack of distribution alignment guidance. When $\boldsymbol{F}_{\text{LSCR}}$ is combined with $\mathcal{L}_{\text{DAMI}}$, the resizer learns domain-invariant features, but it does not fully address the distribution issues. Only by combining $\boldsymbol{F}_{\text{LSCR}}$, $\boldsymbol{F}_{\text{DADS}}$ and $\mathcal{L}_{\text{DAMI}}$ can we alleviate the discrepancy issue, which results in the highest improvement in AP@0.70.

\subsubsection{Contribution of Domain Separation Modules} \label{dads}
We also analyze the effectiveness of the encoder-specific and encoder-agnostic domain separation modules in DADS. As shown in Table~\ref{tab:Ablation_DADS}, using either encoder-agnostic or encoder-specific modules alone does not yield satisfactory results. This is because the encoder-agnostic modules fail to project features into a common space due to interference from encoder-specific information. Similarly, the encoder-specific modules alone cannot completely eliminate domain-dependent information, which may hinder distribution alignment. Furthermore, utilizing two encoder-specific modules does not achieve comparable performance to the combination of encoder-specific and encoder-agnostic modules, indicating the necessity of weight sharing among the second modules. Finally, we examine whether additional encoder-specific modules would help remove domain-dependent information. The results show that a single encoder-specific module is sufficient to remove such information, while additional encoder-specific modules may lead to overfitting and decreased performance. Furthermore, additional encoder-agnostic modules will not improve feature alignment. Because deeper weight-sharing layers over-compress information and over-smooth features, which hampers fine-grained detection.

\begin{table}[htbp]
  \scriptsize
  \renewcommand\arraystretch{1.2} 
  \centering 
  \caption{Ablation study of domain separation modules on V2V4Real.}
  \resizebox{0.8\linewidth}{!}{
      \begin{tabular}{ccccc}
          \toprule
          Encoder  & DADS modules    & AP@0.50   & AP@0.70     \\
          \midrule
          \multirow{6}*{$p_0$+$p_1$}       & $\boldsymbol{M}^{\text{ea}}$           & 2.61    & 1.15    \\
                 & $\boldsymbol{M}^{\text{es}}$           & 58.91    & 35.48    \\
                 & $\boldsymbol{M}^{\text{es}}$+$\boldsymbol{M}^{\text{es}}$      & 48.47    & 27.75    \\
                 & $\boldsymbol{M}^{\text{es}}$+$\boldsymbol{M}^{\text{ea}}$      & \textbf{60.99}    & \textbf{37.89}   \\
                 & 2*$\boldsymbol{M}^{\text{es}}$+$\boldsymbol{M}^{\text{ea}}$    & 58.85    & 31.14   \\
                 & $\boldsymbol{M}^{\text{es}}$+2*$\boldsymbol{M}^{\text{ea}}$   & 0.46 & 36.88   \\
          \midrule
          \multirow{6}*{$p_0$+$s_1$}       & $\boldsymbol{M}^{\text{ea}}$           & 26.03    & 3.67    \\
                 & $\boldsymbol{M}^{\text{es}}$           & 54.23    & 29.88   \\
                 & $\boldsymbol{M}^{\text{es}}$+$\boldsymbol{M}^{\text{es}}$      & 56.91   & 27.68   \\
                 & $\boldsymbol{M}^{\text{es}}$+$\boldsymbol{M}^{\text{ea}}$      & \textbf{71.27}    & \textbf{46.52}    \\
                 & 2*$\boldsymbol{M}^{\text{es}}$+$\boldsymbol{M}^{\text{ea}}$    & 56.52    & 33.86     \\
                 &   $\boldsymbol{M}^{\text{es}}$+2*$\boldsymbol{M}^{\text{ea}}$   & 67.26 & 40.58   \\
          \midrule
          \multirow{6}*{$p_0$+$v_1$}       & $\boldsymbol{M}^{\text{ea}}$           & 47.61    & 11.01   \\
                 & $\boldsymbol{M}^{\text{es}}$           & 62.67    & 39.35     \\
                 & $\boldsymbol{M}^{\text{es}}$+$\boldsymbol{M}^{\text{es}}$      & 58.92    & 31.14   \\
                 & $\boldsymbol{M}^{\text{es}}$+$\boldsymbol{M}^{\text{ea}}$      & \textbf{69.78}    & \textbf{43.41}    \\
                 & 2*$\boldsymbol{M}^{\text{es}}$+$\boldsymbol{M}^{\text{ea}}$    & 66.62    & 34.37   \\
                 & $\boldsymbol{M}^{\text{es}}$+2*$\boldsymbol{M}^{\text{ea}}$    & 67.02 & 36.01  \\
          \bottomrule
          \end{tabular}
  }
  \label{tab:Ablation_DADS}
  \vspace{-1em}
\end{table}

\section{Conclusion}

In this paper, we propose CoDS, a fully convolutional collaborative perception adapter to address feature discrepancies in heterogeneous scenarios through domain separation. Specifically, the CoDS incorporates the LSCR to align feature dimensions, followed by the DADS module, which removes encoder-specific information while preserving task-relevant information. During training, CoDS employs DAMI loss to further enhance the domain separation process.
Extensive experiments on the V2V4Real, OPV2V and V2XSet datasets demonstrate that CoDS effectively mitigates feature discrepancies and consistently achieves an optimal balance between detection accuracy and inference efficiency.

\bibliographystyle{IEEEtran}
\bibliography{references}


\vspace{11pt}

\vspace{11pt}


\vfill

\end{document}